\documentclass[journal,10pt]{IEEEtran}
\usepackage{amsmath,amsfonts}
\usepackage{algorithmic}
\usepackage{algorithm}
\usepackage{array}
\usepackage[caption=false,font=normalsize,labelfont=sf,textfont=sf]{subfig}
\usepackage{textcomp}
\usepackage{stfloats}
\usepackage{url}
\usepackage{verbatim}
\usepackage{graphicx}
\usepackage{cite}
\usepackage{booktabs}
\usepackage{multirow}
\usepackage{amssymb}
\usepackage[sort,numbers]{natbib}
\usepackage{threeparttable}
\usepackage{hyperref}
\usepackage{color} 
\usepackage{marvosym} 
\usepackage{hyperref}[colorlinks,linkcolor=blue] 

\begin{document}

\title{Collaborative Perception in Autonomous Driving: Methods, Datasets and Challenges}

\author{Yushan Han, Hui Zhang,~\IEEEmembership{Member,~IEEE,} Huifang Li, \\ Yi Jin, ~\IEEEmembership{Member,~IEEE,} Congyan Lang,  Yidong Li\textsuperscript{\Letter}, ~\IEEEmembership{Senior Member,~IEEE}
\thanks{This work was supported by the National Natural Science Foundation of China under Grant U1934220 and U2268203.}
\thanks{The authors are with Key Laboratory of Big Data \& Artificial Intelligence in Transportation, Ministry of Education (Beijing Jiaotong University) and School of Computer and Information Technology, Beijing Jiaotong University, Beijing 100044, China.
(Email: yushanhan@bjtu.edu.cn; huizhang1@bjtu.edu.cn; 17112084@bjtu.edu.cn; yjin@bjtu.edu.cn; cylang@bjtu.edu.cn; ydli@bjtu.edu.cn)}
\thanks{Corresponding author: Yidong Li.}
\thanks{Project Page: \href{https://github.com/CatOneTwo/Collaborative-Perception-in-Autonomous-Driving}{https://github.com/CatOneTwo/Collaborative-Perception-in-Autonomous-Driving}}
\thanks{Manuscript received XXX, XX, 2023; revised XXX, XX, 2023.}}

\markboth{IEEE Intelligent Transportation Systems Magazine, ~Vol.~XX, No.~XX, XXX~2023}
{}

\maketitle

\begin{abstract}
  Collaborative perception is essential to address occlusion and sensor failure issues in autonomous driving. In recent years, theoretical and experimental investigations of novel works for collaborative perception have increased tremendously. So far, however, few reviews have focused on systematical collaboration modules and large-scale collaborative perception datasets. This work reviews recent achievements in this field to bridge this gap and motivate future research. We start with a brief overview of collaboration schemes. After that, we systematically summarize the collaborative perception methods for ideal scenarios and real-world issues. The former focuses on collaboration modules and efficiency, and the latter is devoted to addressing the problems in actual application. Furthermore, we present large-scale public datasets and summarize quantitative results on these benchmarks. Finally, we highlight gaps and overlook challenges between current academic research and real-world applications.
\end{abstract}


\IEEEpeerreviewmaketitle

\section{Introduction}
\IEEEPARstart{A}{utonomous} driving is a prominent technology in research and commercial vehicles \cite{schwarz2022role,shi2022integrated,9043473}.
From a broad perspective, an autonomous driving system contains perception, planning and control modules \cite{av}.
The perception module utilizes sensors to continuously scan and monitor the surroundings, which is vital for autonomous vehicles (AV) to understand environments.
AV perception can be divided into individual perception and collaborative perception. 
Although individual perception has made significant progress with the development of deep learning \cite{zhang2019mask,zhang2021c2fda,vsln,zhang2017advances,li2022features, li2022novel}, some problems limit its development.
Firstly, individual perception often encounters occlusion when perceiving a relatively comprehensive environment. Secondly, onboard sensors have physical limitations in sensing distant objects. Furthermore, sensor noise degrades the performance of the perception system.

\begin{figure}
    \centering
    \includegraphics[width=1\linewidth]{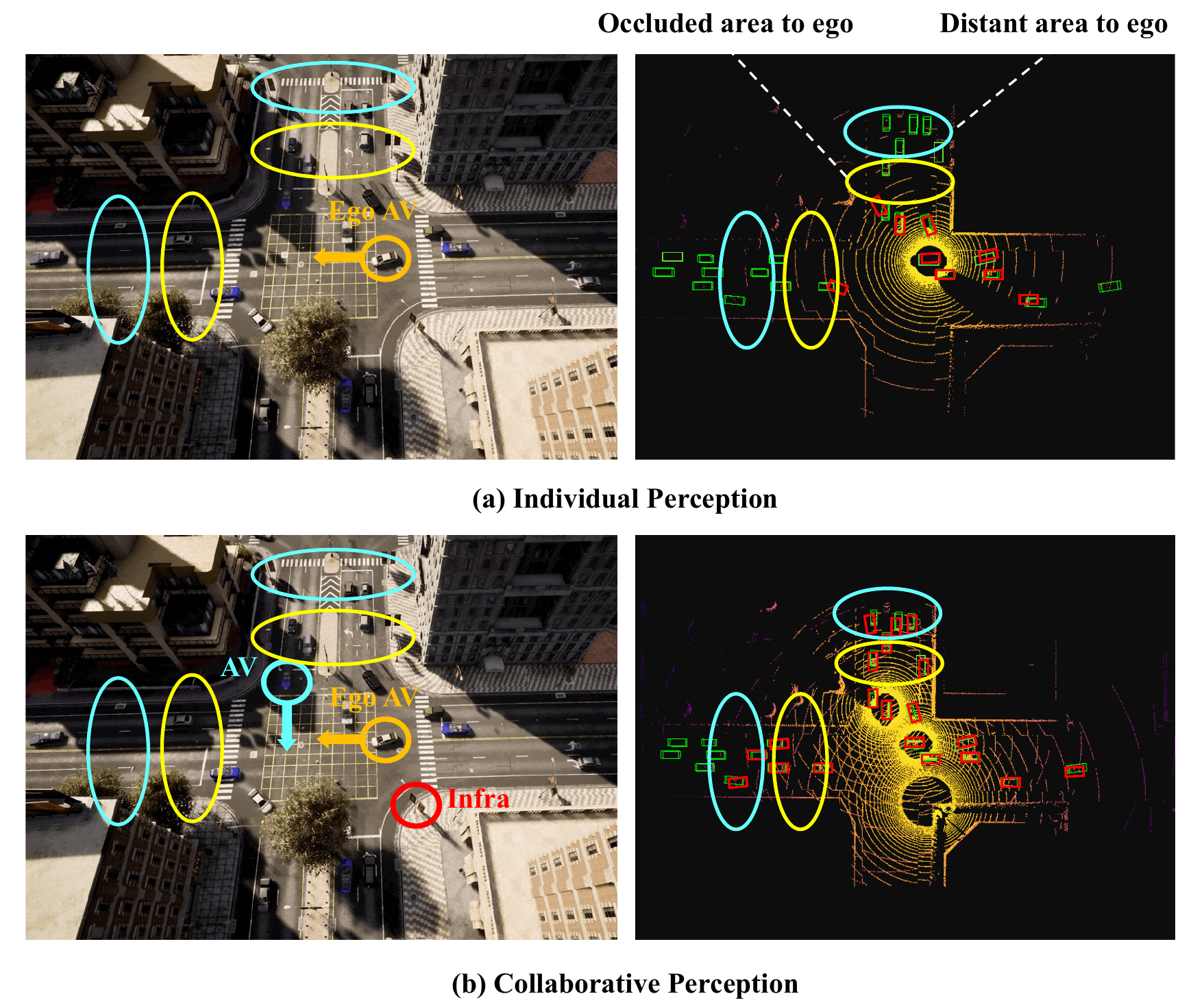}
    \caption{An example of (a) individual perception and (b) collaborative perception in autonomous driving. \textit{Left}: Screenshot of one autonomous driving scenario. \textit{Right}: Point cloud schematic. The green and red bounding boxes represent ground truths and predictions respectively. The yellow and blue ellipses represent occluded and distant areas of the ego vehicle. Ego AV: ego autonomous vehicle, AV: other autonomous vehicles, Infra: infrastructures.}
    \label{example}
    \vspace{-1.2em}
\end{figure}

To compensate for deficiencies in individual perception, collaborative or cooperative perception, which exploits the interaction among multiple agents, has received considerable attention. 
Collaborative perception is a multi-agent system \cite{mas} in which agents share perceptual information to overcome visual limitations in ego AV. 
As shown in Fig. \ref{example}, in an individual perception scenario, the ego AV only detects a part of nearby objects for occlusion and sparse point clouds in distant areas.
In a collaborative perception scenario, the ego AV expands the field of view by receiving information from other agents.
Through this collaboration way, the ego AV not only detects distant and occluded objects but also improves the detection accuracy in dense areas.

Collaborative perception has been in the spotlight for a long time.
Previous works \cite{rauch2012car2x,kim2014multivehicle, ZHAO2017407,rawashdeh2018collaborative, Collaborativev2v} have focused on building collaborative perception systems to evaluate the feasibility of this technology. 
However, it has not been effectively advanced due to the lack of large public datasets.
In recent years, there has been a surge in interest and research with the development of deep learning and the public of large-scale collaborative perception datasets \cite{OPV2V, V2X-Sim, DAIR-V2X}. 
Considering bandwidth constraints in communication, most researchers \cite{v2vnet, DiscoNet,where2comm} are devoted to designing novel collaboration modules to achieve a trade-off between accuracy and bandwidth. 
However, the above works assume a perfect collaborative scenario.
To alleviate some issues in practical autonomous driving applications, such as localization errors, communication latency and model discrepancy, recent works \cite{RobustV2VNet, SyncNet} propose corresponding solutions to ensure the robustness and safety of the collaborative system.

\begin{figure*}
    \centering
    \includegraphics[width=1\linewidth]{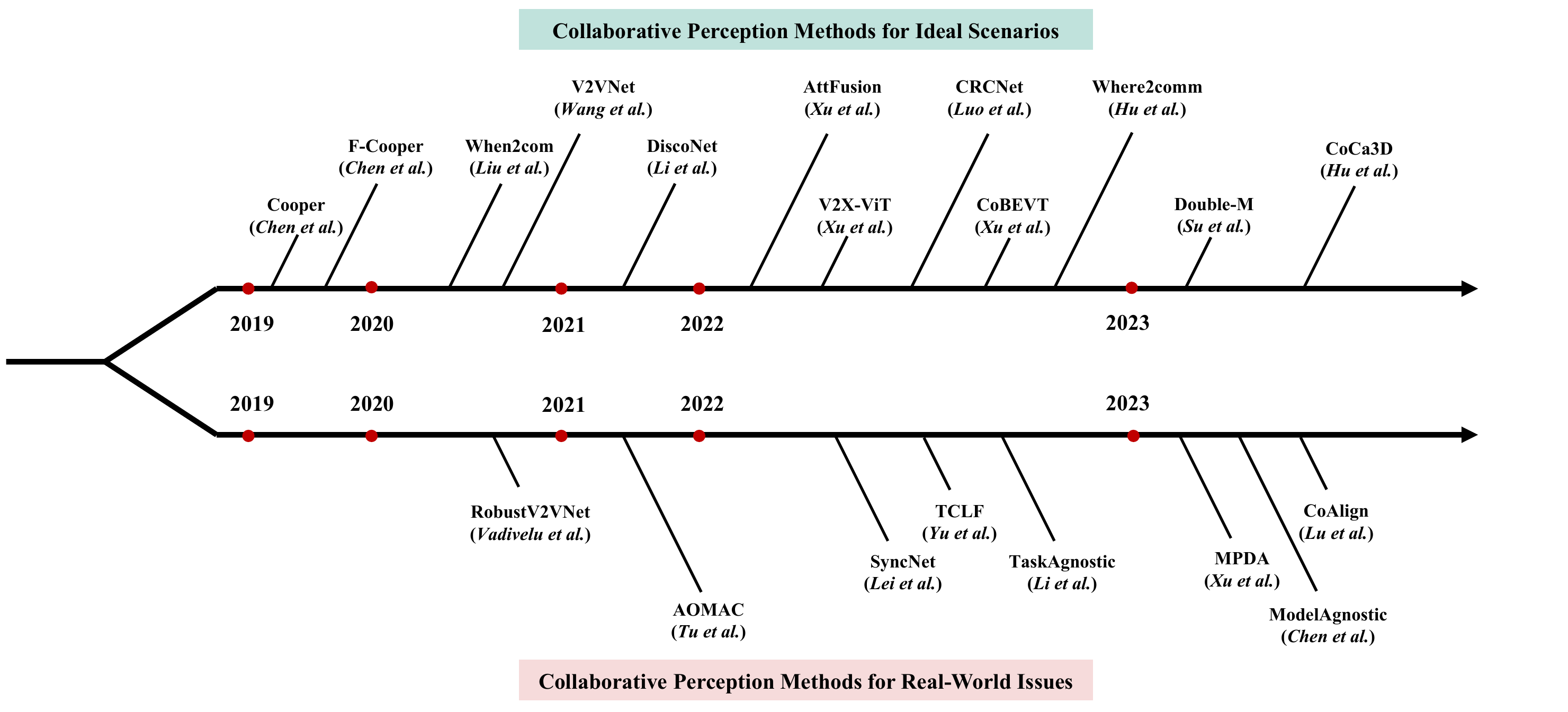}
    \caption{
        Typical collaborative perception methods in autonomous driving are classified from two perspectives:
        1) how to design common collaboration modules in ideal scenarios, which focus on collaboration efficiency and performance, and 2) how to address issues in real applications, which focus on robustness and safety. We categorized methods based on their most prominent contribution. Citation: 1) Cooper \cite{cooper}, F-Cooper \cite{fcooper}, When2com \cite{when2com}, V2VNet \cite{v2vnet}, DiscoNet \cite{DiscoNet}, AttFusion \cite{OPV2V}, V2X-ViT \cite{V2X-VIT}, CRCNet \cite{CRCNet}, CoBEVT \cite{CoBEVT}, Where2comm \cite{where2comm}, Double-M \cite{su2022uncertainty}, CoCa3D \cite{hu2023collaboration}, 2) RobustV2VNet \cite{RobustV2VNet}, AOMAC \cite{AOMAC}, SyncNet \cite{SyncNet}, TCLF \cite{DAIR-V2X}, TaskAgnostic \cite{csc}, MPDA \cite{xu2022bridging}, ModelAgnostic \cite{PSA}, CoAlign \cite{coalign}.}
    \label{fig1:milestones}
\end{figure*}

To summarize these technologies and issues, we go over collaborative perception methods in autonomous driving and give a comprehensive survey of recent advances in terms of methods, datasets and challenges. 
We also notice some reviews on collaborative perception \cite{caillot2022survey,ren2022collaborative,cui2022cooperative} have been published in recent years.
The main difference between this paper and existing reviews are summarized as follows:

\begin{itemize}

    \item Firstly, most of the previous reviews merely focus on some specific application issue \cite{cui2022cooperative} or perception task \cite{caillot2022survey}. In this work, we provide a systematic summary of collaboration methods, which will help readers to establish a complete knowledge system and find the future direction rapidly. Specifically, we review recent works on collaboration modules in ideal scenarios and solutions for real-world issues.
    The former pays attention to collaboration efficiency and performance, while the latter focuses more on collaboration robustness and safety, as shown in Fig. \ref{fig1:milestones}.

    \item Secondly, although current reviews have discussed some previous methods, they didn't cover the latest research progress, such as new application problems, state-of-the-art frameworks and large public datasets. 
    To this end, we track and summarize these latest developments.
    As far as we know, this is the first work to summarize and compare large-scale collaborative perception datasets comprehensively. Furthermore, we propose challenges and opportunities for future work by discussing the current research state.
\end{itemize}

This paper is structured as follows. 
Section \uppercase\expandafter{\romannumeral2} briefly presents the collaboration scheme.
Section \uppercase\expandafter{\romannumeral3} systematically analyzes the collaboration methods in autonomous driving perception networks.
We summarize well-referenced large-scale collaborative perception datasets and compare the performance of existing methods in Section \uppercase\expandafter{\romannumeral4}.
Open challenges and promising directions are discussed in Section \uppercase\expandafter{\romannumeral5}.
Finally, section \uppercase\expandafter{\romannumeral6} provides a summary and concludes this work.
\section{Collaboration scheme}

\begin{figure*}[htbp]
    \centering
    \includegraphics[width=\linewidth,scale=1.00]{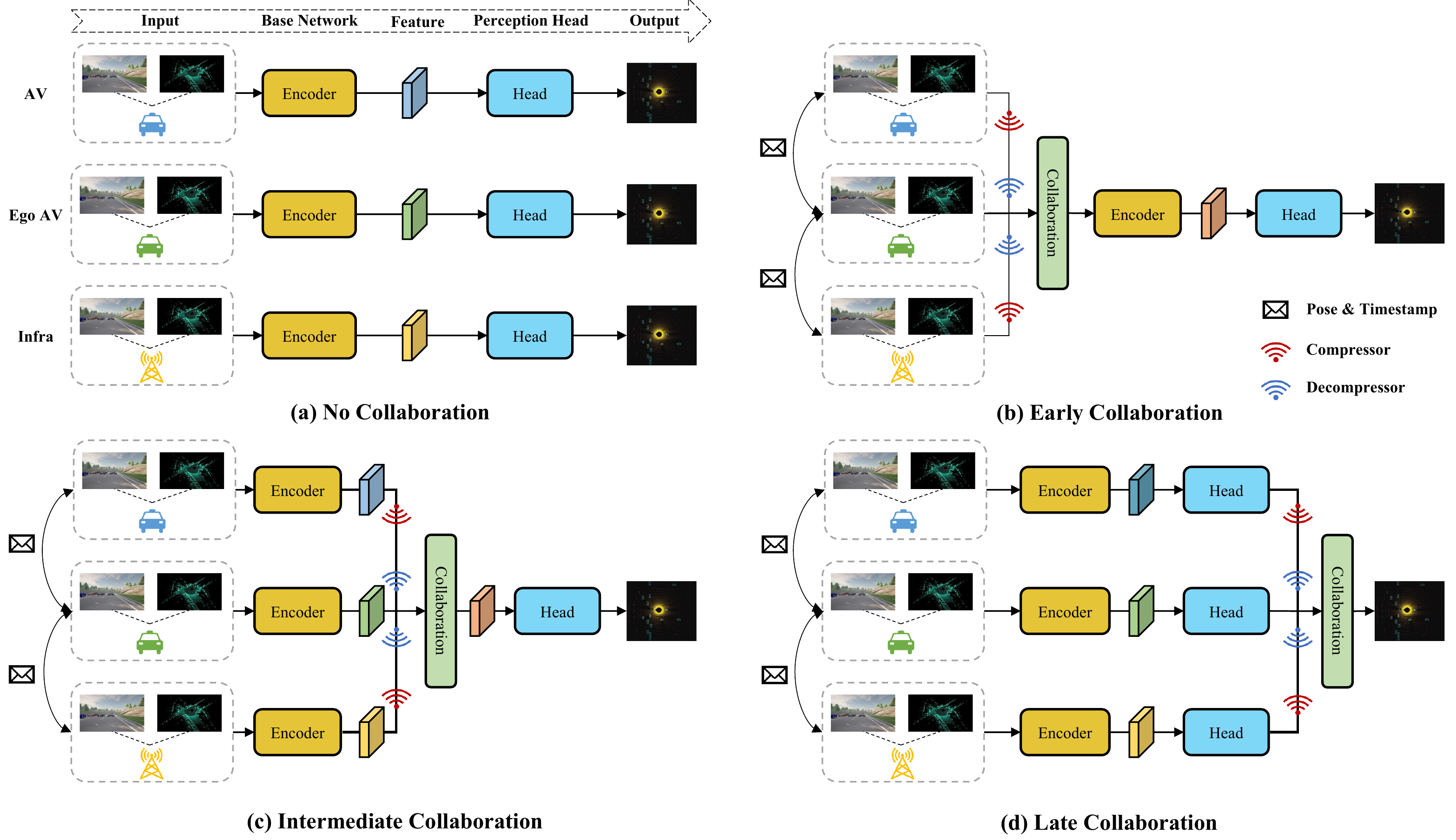}
    \caption{The collaboration scheme in the collaborative perception system. (a) shows the framework of individual perception or no collaboration. (b-d) demonstrate three general frameworks of collaborative perception in autonomous driving. Early collaboration (b) transmits and fuses raw data at the input of the perception network, intermediate collaboration (c) aggregates features, and late collaboration (d) merges outputs directly.}
    \label{fig3:fusion_scheme}
\end{figure*}

The general AV perception system contains input, base network, feature, perception head and output stages.
According to the data sharing and collaboration stage, the collaborative perception scheme can be broadly separated into early, intermediate and late collaboration, as shown in Fig. \ref{fig3:fusion_scheme} (b-d). In order to facilitate readers understanding the relationship between individual perception and the three collaborative schemes, we also illustrate individual perception in Fig. \ref{fig3:fusion_scheme} (a). 

\subsection{Early Collaboration}

Early collaboration employs the raw data fusion at the input of the network, which is also known as data-level or low-level fusion (Fig. \ref{fig3:fusion_scheme} (b)). In an autonomous driving scene, the ego vehicle receives and transforms the raw sensor data from other agents, and then aggregates transformed data onboard. 
Raw data contains the most comprehensive information and substantial description of agents. Consequently, early collaboration can fundamentally overcome occlusion and long-range problems in individual perception and promote performance to the greatest extent. However, early collaboration relies on high data bandwidth, which makes it challenging to achieve real-time edge computing.

\subsection{Intermediate Collaboration}
Considering the high bandwidth of early collaboration, some works propose intermediate collaborative perception methods to balance the performance-bandwidth trade-off. In intermediate collaboration (Fig. \ref{fig3:fusion_scheme} (c)), other agents usually transfer deep semantic features to the ego vehicle. The ego vehicle fuses features to make the ultimate prediction.
Intermediate collaboration has become the most popular multi-agent collaborative perception choice for flexibility.
However, feature extraction often causes information loss and unnecessary information redundancy, which motivates people to explore suitable feature selection and fusion strategies. 

\subsection{Late Collaboration}

Late or object-level collaboration employs the prediction fusion at the network output, as shown in Fig. \ref{fig3:fusion_scheme} (d). Each agent trains the network individually and shares outputs with each other. The ego vehicle spatially transforms the outputs and merges all outputs after postprocessing. 
Late collaboration is more bandwidth-economic and simpler than early and intermediate collaboration. However, the late collaboration also has limitations. Since individual output could be noisy and incomplete, late collaboration always has the worst perception performance.
\section{Collaborative perception methods}
Many collaborative perception methods have emerged recently.
Since these methods are based on different frameworks and objectives, it's urgent to establish a systematic taxonomy to help readers understand this field.
To this end, this section reviews recent collaborative perception approaches systematically.
Specifically, we review collaborative perception from the aspect of methods for ideal autonomous driving scenarios (Sec. \ref{m2}), and for issues in real applications along with their solutions (Sec. \ref{robustness}).
The former pays attention to collaboration efficiency and performance, while the latter focuses more on collaboration robustness and safety.
The reviewed methods involve V2V (vehicle-to-vehicle), V2I (vehicle-to-infrastructure) and V2X (vehicle-to-everything) modes, and detailed summaries of recent progress are provided in Tab. \ref{table:methods1} and Tab. \ref{table:methods2}.

\subsection{Methods for Ideal Scenarios} \label{m2}

\begin{figure*}[htbp]
    \centering
    \includegraphics[width=0.8\linewidth,scale=1.00]{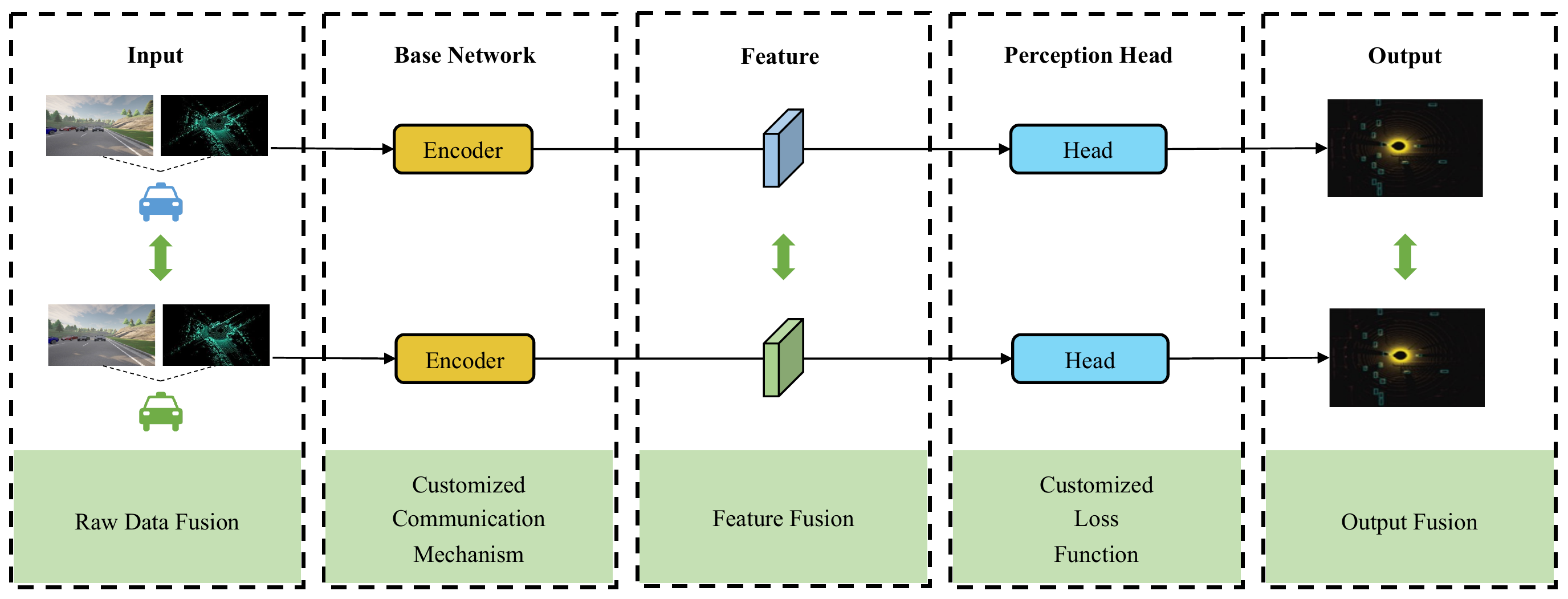}
    \caption{Illustration of the common collaboration modules present in collaborative perception networks, which are used to improve collaboration efficiency and performance. Common collaboration modules include raw data fusion, customized communication mechanism, feature fusion, customized loss function and output fusion. Various approaches will design appropriate collaboration strategies based on the collaboration scheme and objectives.
    }
    \label{fig2:collaboration strategy}
\end{figure*}

\begin{figure*}[htbp]
    \centering
    \includegraphics[width=1.0\linewidth,scale=1.00]{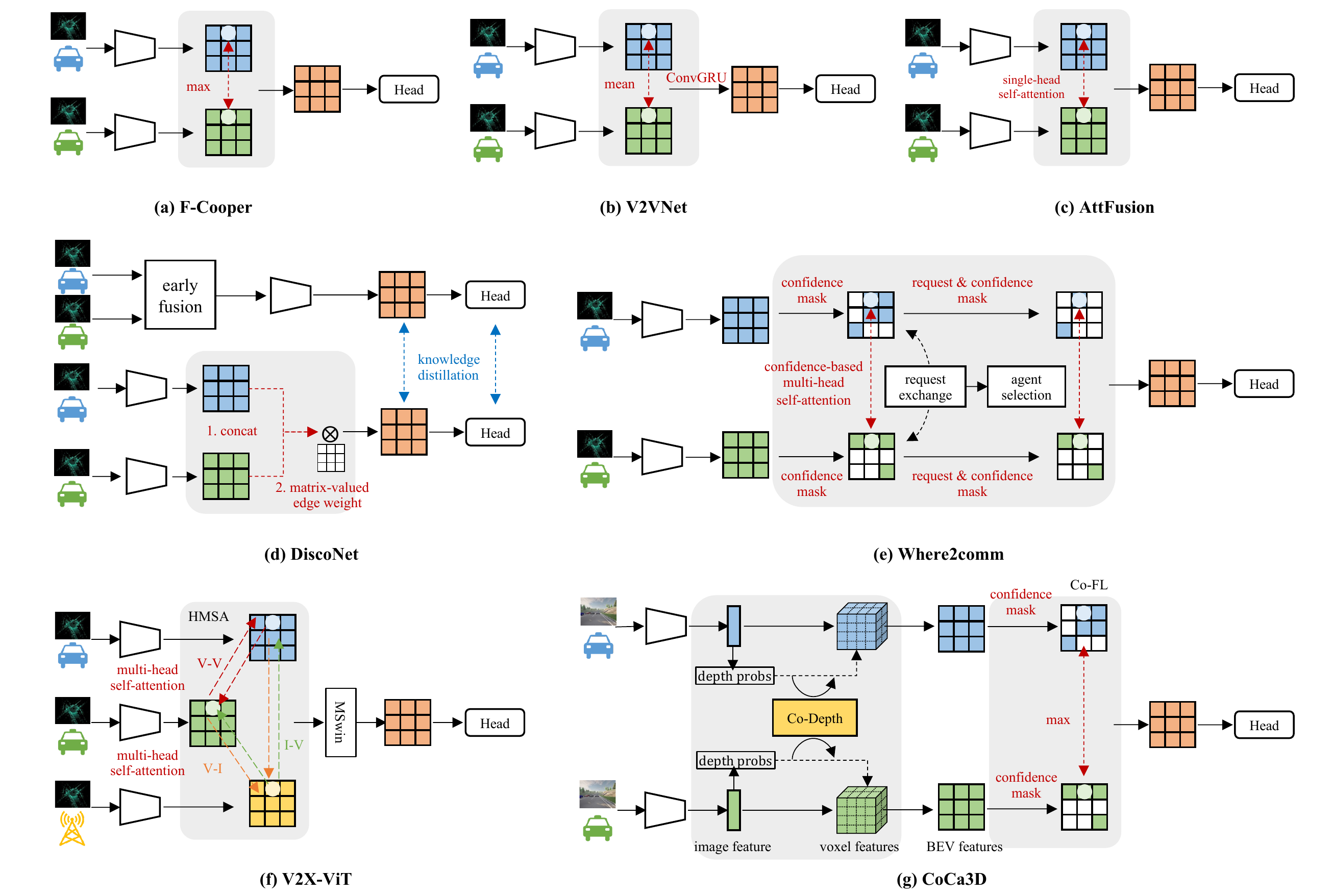}
    \caption{Illustration of state-of-the-art intermediate collaborative perception methods in ideal scenarios, and we focus on feature fusion modules here.\\ (a) F-Cooper \cite{fcooper}, (b) V2VNet \cite{v2vnet}, (c) AttFusion \cite{OPV2V}, (d) DiscoNet \cite{DiscoNet}, (e) Where2comm \cite{where2comm}, (f) V2X-ViT \cite{V2X-VIT}, (g) CoCa3D \cite{hu2023collaboration}.
    }
    \label{fig2:collaboration method}
\end{figure*}

In collaborative perception systems, researchers usually design corresponding collaboration modules at different network stages to improve collaborative efficiency and perception performance.
Generally, three collaboration schemes require basic collaboration modules to aggregate multi-agent observations, such as raw data fusion at the input stage (early collaboration), feature fusion at the feature processing stage (intermediate collaboration), and output fusion at the output stage (late collaboration). Furthermore, some works establish communication mechanisms in the base network stage to reduce transmission bandwidth, and introduce customized loss functions at the perception head stage to guide the network to capture helpful information. 
In this subsection, we summarize and introduce the basic collaboration modules of each stage separately, as shown in Fig. \ref{fig2:collaboration strategy}. In addition, an illustration of the state-of-the-art intermediate collaborative perception framework is provided in Fig. \ref{fig2:collaboration method}

\subsubsection{Raw Data Fusion}

Early collaboration adopts raw data fusion at the input stage. Since point clouds are irregular and can be aggregated directly, early collaboration works \cite{cooper, Coop3D} usually adopt point cloud fusion strategies.

The first early collaborative perception system Cooper \cite{cooper} chooses LiDAR data as a fusion target. Point clouds can be compressed into smaller sizes by only extracting positional coordinates and reflection values. 
After interaction among agents, Cooper reconstructs the received point clouds with a transformation matrix, and then concatenates the ego point clouds set to make final prediction.


Inspired by Cooper, Coop3D \cite{Coop3D} also explores early collaboration and introduces a new point cloud fusion method. Specifically, instead of employing concatenation, the Coop3D system utilizes the spatial transformation to fuse sensor data.
Besides, unlike Cooper sharing vehicle-to-vehicle information on board, Coop3D proposes a central system to merge multiple sensor data, which allows the amortization of sensor and processing costs in collaboration.

\subsubsection{Customized Communication Mechanism}

Raw data fusion in early collaboration broadens the horizons of the ego vehicle, which also causes high bandwidth pressure. 
To alleviate the above issues, more and more works \cite{v2vnet, DiscoNet, V2X-VIT} develop intermediate collaboration.
The initial intermediate collaboration methods follow a greedy communication mechanism to obtain as much information as possible.
Generally, they share information with all agents within the communication range and put the compressed full feature maps into collective perception messages (CPMs). However, since the feature sparsity and agent redundancy, the greedy communication may waste the bandwidth hugely. 
To fill this gap, some works \cite{who2com, FPV-RCNN,where2comm, Learn2com} establish dynamic communication mechanisms to select agents and features.

Who2com \cite{who2com} establishes the first communication mechanism under bandwidth constraints, which is realized via a three-stage handshake. Specifically, Who2com uses a general attention function \cite{generalatt} to calculate match scores among agents and selects the most needed agents to reduce bandwidth effectively. Based on Who2com, When2com \cite{when2com} introduces the scaled general attention to determine when to communicate with others. Thus, the ego agent communicates with others only when information is insufficient, effectively saving collaboration resources.

In addition to suitable communication agent selection, communication content is important to reduce bandwidth pressure.
The initial feature selection strategy is proposed in FPV-RCNN \cite{FPV-RCNN}. 
Specifically, FPV-RCNN adopts a detection head to generate proposals and only selects the feature points in the proposals. The key points selection module reduces the redundancy of shared deep features and provides valuable supplementary information to initial proposals.

Where2comm \cite{where2comm} also proposes a novel spatial-confidence-aware communication mechanism. Its core idea is to utilize a spatial confidence map to decide the sharing features and communication targets. In the feature selection stage, Where2comm selects and transmits spatial elements that satisfy high confidence and other agents' request. In the agent selection stage, the ego agent only communicates with agents who can provide the required features. By sending and receiving features on perceptually critical areas, Where2comm saves massive bandwidth and significantly improves collaboration efficiency.

\subsubsection{Feature Fusion}

Feature fusion module is crucial in intermediate collaboration. After receiving CPMs from other agents, the ego vehicle can leverage different strategies to aggregate these features. 
A feasible fusion strategy is able to capture latent relationships between features and improve the performance of the perception network.
According to the idea of feature fusion based on, we divided the existing feature fusion methods into traditional, graph-based, and attention-based fusion.

\paragraph{Traditional Fusion}

At the early stage of the research on collaborative perception, researchers tend to use traditional strategies to fuse features, such as concatenation, sum and linear weighted. Intermediate collaboration applies these invariant permutation operations on deep features \cite{fcooper, CoFF}, which achieves fast inference due to simplicity.

The first intermediate collaborative perception framework F-Cooper \cite{fcooper} extracts low-level voxel and deep spatial features. Based on these two level features, F-Cooper proposes two feature-fusion strategies: Voxel Feature Fusion (VFF) and Spatial Feature Fusion (SFF). Both employ element-wise maxout to fuse features in overlapped regions. 
Since voxel features are closer to raw data, VFF is as capable as the raw data fusion method for near object detection. In the meanwhile, SFF also has its advantages. Inspired by SENet \cite{senet}, SFF opts to select partial channels to reduce transmission time consumption while keeping the comparable detection precision.

Considering F-Cooper \cite{fcooper} ignores the importance of low-confidence features, Guo et al. \cite{CoFF} propose CoFF to improve F-Cooper. CoFF weights the overlapped features by measuring their similarity and overlapping area. The smaller the similarity and the greater the distance, the more supplementary information provided by neighbor features intuitively. Besides, an enhancement parameter is added to increase the value of weak features. Experiments show that the simple yet efficient design makes CoFF improve F-Cooper much.

Although simple, traditional fusion has not been abandoned by recent methods. Hu et al. \cite{hu2023collaboration} propose collaborative camera-only 3D detection (CoCa3D) to demonstrate the potential of collaboration in enhancing camera-based 3D detection. Since depth estimation is the bottleneck of camera-based 3D detection, CoCa3D contains collaborative depth estimation (Co-Depth) except for collaborative feature learning (Co-FL). In Co-Depth, neighbour agents transmit depth estimation with low uncertainty only, and the ego agent updates the depth estimation by considering single-view depth probability and multi-view consistency. In Co-FL, agents send feature elements with high detection confidence and adopt a simple non-parametric point-wise maximum to fuse features. Experimental results indicate that CoCa3D helps the camera overtake LiDAR in 3D object detection.

\paragraph{Graph-based Fusion}

Despite the simplicity of traditional intermediate fusion, they ignore the latent relationship among multi-agents and fail to reason the messages from sender to receiver.
Graph Neural Networks (GNNs) have the ability to propagate and aggregate messages from neighbors \cite{gnn}, and recent works have shown the effectiveness of GNN on perception and autonomous driving.

V2VNet \cite{v2vnet} firstly leverages a spatial-aware graph neural network (GNN) to model the communication among agents. 
In GNN message passing stage, V2VNet utilizes a variational image compression algorithm to compress features. In cross-vehicle aggregation, V2VNet first compensates for the time delay to create an initial state for each node and then warps and spatially transforms the compressed features from neighbor agents to the ego vehicle, all these operations are conducted in overlapping fields of view. In the feature fusion stage, V2VNet adopts average operation to aggregate features and updates the node state with a convolutional gated recurrent unit (ConvGRU). 

Although V2VNet \cite{v2vnet} achieves performance improvement with GNN, the scalar-valued collaboration weight cannot reflect the importance of the different spatial regions. Motivated by this, DiscoNet \cite{DiscoNet} proposes to use matrix-valued edge weight to capture the inter-agent attention in high resolution. During the message passing, DiscoNet concatenates features and applies a matrix-valued edge weight for each element in feature maps. Besides, DiscoNet combines early and intermediate fusion by applying a teacher-student framework, which further improves performance. 

Zhou et al. \cite{MP-Pose} propose another generalized GNN-based perception framework MP-Pose. During the message passing stage, MP-Pose encodes the relative spatial relationship with a spatial encoding network rather than warps features directly \cite{v2vnet, DiscoNet}. Inspired by Graph Attention Networks (GAT), it further uses a dynamic cross-attention encoding network to capture the relationship among agents and aggregate multiple features like GAT.

\paragraph{Attention-based Fusion}

In addition to graph learning, attention mechanisms have emerged as a powerful tool for exploring feature relationships \cite{attentionsurvey, transformer}. Attention mechanisms can be classified according to the data domain into channel attention, spatial attention, and channel \& spatial attention \cite{attentionsurvey}. Over the past decade, attention mechanisms have played an increasingly important role in computer vision \cite{senet,vit,detr} and have inspired collaborative perception research. Since feature selection and relationship exploring are vital issues in intermediate collaborative perception, some works have leveraged attention mechanisms \cite{who2com, when2com, FAR, OPV2V, V2X-VIT, SyncNet, CRCNet} to develop more dynamic and robust feature fusion strategies. Due to flexibility, attention-based designs have dominated intermediate collaboration.

To capture the interaction among the specific area in the feature maps, Xu et al. \cite{OPV2V} propose AttFusion and first employ a self-attention operation at the exact spatial location. Specifically, AttFusion introduces a single-head self-attention fusion module and achieves a balance between performance and inference speed compared to the traditional method F-Cooper \cite{fcooper} and graph-based method DiscoNet \cite{DiscoNet}. The spatial-aware interaction in AttFusion is similar to the matrix-weight edge in DiscoNet \cite{DiscoNet} while implemented with different tools.

Besides traditional attention-based methods, Transformer-based methods also inspire collaborative perception. Cui et al. \cite{COOPERNAUT} propose COOPERNAUT based on Point Transformer \cite{pointtransformer}, a self-attention network for point cloud processing. After receiving messages, the ego agent uses a down-sampling block and point transformer block to aggregate points features. The former block is used to reduce the cardinality of the point sets, and the second block allows local information exchange among all points. Both two operations preserve the permutation invariance of messages. What is more important, COOPERNAUT integrates collaborative perception with control decisions, which is of great significance for the module linkage of autonomous driving.

Compared with V2V collaboration, V2I could provide more stable collaboration information with a tremendous amount of infrastructures, whereas there have few works have paid attention to this scenario. Xu et al. \cite{V2X-VIT} present the first unified transformer architecture (V2X-ViT), which covers V2V and V2I simultaneously. To module interactions among different types of agents, V2X-ViT proposes a novel heterogeneous multi-agent attention module (HMSA) to learn the different relationships between V2V and V2I. Furthermore, a multi-scale window attention module (MSwin) is introduced to capture long-range spatial interaction on high-resolution detection. 

Furthermore, considering RGB camera is cheaper than LiDAR, Xu et al. \cite{CoBEVT} present the first generic multi-camera-based collaborative perception framework CoBEVT. CoBEVT designs a fused axial attention (FAX) module to explore the interaction among multi-view and multiple agents by performing sparse global interactions and local window-based attention.
Experiments demonstrate that CoBEVT performs well in multi-view and multi-agent interaction, with robustness to camera loss.

Wang et al. \cite{wang2023vimi} also introduce a camera-based collaborative perception method called the vehicles-infrastructure multi-view intermediate fusion (VIMI). To explore the correlation between features of vehicles and infrastructure at different scales, VIMI employs a multi-scale cross-attention (MCA) module that extracts multi-scale features using deformable CNN and generates attention weights for each scale through the cross-attention operation. Additionally, they design a camera-aware channel masking strategy (CCM), which corrects calibration errors and augments features by re-weighting the features in a channel-wise manner based on camera parameters.

Previous studies have primarily focused on homogeneous sensor collaboration scenarios. However, collaborative perception involving heterogeneous sensors has not yet been explored. To fill this gap, Xiang et al. propose the first unified hetero-modal vehicle-to-vehicle (V2V) cooperative perception framework HM-ViT \cite{HM-ViT}. HM-ViT comprises a generic heterogeneous 3D graph attention (H3GAT) that fuses bird's eye view (BEV) features from distinct sensor types in multiple agents. Specifically, type-dependent nodes and edges in H3GAT jointly capture sensor heterogeneity, reason spatial interactions and cross-agent relations. Additionally, local window-based attention and sparse global grid-based attention are introduced to capture local and global cues. Experimental results demonstrate the superiority of HM-ViT in both hetero-modal and homo-modal collaborative perception.

\begin{table*}[]
    \vspace{-1em}
    \renewcommand\arraystretch{1.3} 
    \centering 
    \caption{A summary of state-of-the-art collaborative perception methods for ideal scenarios.}
    \begin{threeparttable}
    \resizebox{\linewidth}{!}{ 
    \begin{tabular}{cc|cc|cccc|c}
    \toprule
    \textbf{Method}       & \textbf{Venue} & \textbf{Modality} &  \textbf{Scheme} & \textbf{Data Fusion} &\textbf{Comm Mecha} & \textbf{Feat Fusion} & \textbf{Loss Func} & \textbf{Code}                                   \\
    \midrule
    \midrule
    Cooper \cite{cooper}    & ICDCS'19  & LiDAR        & E    &Raw              & -     & -        &-&   -    \\
    F-Cooper \cite{fcooper}     & SEC'19      &  LiDAR    & I     &-      & -& Trad              &-&  \href{https://github.com/Aug583/F-COOPER}{Link}                                           \\
    \midrule
    \midrule
    Who2com \cite{who2com}     & ICRA'20          &     Camera           & I     &-       & Agent &Trad          &-&    -   \\
    When2com \cite{when2com}       & CVPR'20     &    Camera   & I        &-    & Agent & Trad          &-&  \href{https://github.com/GT-RIPL/MultiAgentPerception}{Link} \\
    V2VNet \cite{v2vnet}        & ECCV'20    &  LiDAR    & I      &-      & -&Graph              &-&  - \\
    Coop3D \cite{Coop3D}     & TITS'20    &      LiDAR     & E, L      &Raw, Out      & -&-         &- &   \href{https://github.com/eduardohenriquearnold/coop-3dod-infra}{Link}    \\
    \midrule
    \midrule
    CoFF \cite{CoFF}  & IoT'21    &     LiDAR     & I     &-       & -&Trad    &- &     -   \\
    DiscoNet \cite{DiscoNet}    & NeurIPS'21     &    LiDAR    & I      &Raw      & -&Graph       &- &  \href{https://github.com/ai4ce/DiscoNet}{Link}       \\
    \midrule
    \midrule
    MP-Pose \cite{MP-Pose}     & RAL'22    &    Camera     & I &-   & -&Graph    &- &  -       \\
    FPV-RCNN \cite{FPV-RCNN}    & RAL'22   &  LiDAR   & I   &Out   & Feat&Trad    & - & \href{https://github.com/YuanYunshuang/FPV_RCNN}{Link}                                   \\
    AttFusion \cite{OPV2V}  & ICRA'22    &   LiDAR  & I   &-    & -&Atten   &-      & \href{https://github.com/DerrickXuNu/OpenCOOD}{Link}                                      \\
    TCLF \cite{DAIR-V2X}      & CVPR'22    &   LiDAR  & L    &Out   & -&-         &- &   \href{https://github.com/AIR-THU/DAIR-V2X}{Link}                                      \\
    COOPERNAUT \cite{COOPERNAUT} & CVPR'22   & LiDAR   & I      &-      & - &Atten     & - &   \href{https://github.com/UT-Austin-RPL/Coopernaut}{Link}                                  \\
    V2X-ViT \cite{V2X-VIT}   & ECCV'22   &   LiDAR   & I       &-     & -& Atten   & - & \href{https://github.com/DerrickXuNu/v2x-vit}{Link}                                       \\
    CRCNet \cite{CRCNet}     & MM'22    &    LiDAR    & I      &-      & - & Atten & Redund   &      -    \\
    CoBEVT \cite{CoBEVT}    & CoRL'22     &     Camera     & I  &-  & - &Atten     &- &     \href{https://github.com/DerrickXuNu/CoBEVT}{Link} \\
    Where2comm \cite{where2comm}   & NeurIPS'22     &  LiDAR  & I    &-     & Agent, Feat& Atten   & - &    \href{https://github.com/MediaBrain-SJTU/Where2comm}{Link} \\
    \midrule
    \midrule
    Double-M \cite{su2022uncertainty}  & ICRA'23   &    LiDAR  &   E, I, L     &-     &   - & - &Uncert &     \href{https://github.com/coperception/double-m-quantification}{Link}    \\
    CoCa3D \cite{hu2023collaboration}  & CVPR'23   &    Camera  &   I     &-     &   Feat & Trad &- &     \href{https://github.com/MediaBrain-SJTU/CoCa3D}{Link}    \\
    VIMI \cite{wang2023vimi}  & arXiv'23   &      Camera    & I      &-     & - &Atten    & - &       -    \\ 
    HM-ViT \cite{HM-ViT}  & arXiv'23   &   LiDAR, Camera    & I        &-     & - &Atten   & - &       -    \\ 
    \bottomrule
    \end{tabular}
    }
    \begin{tablenotes}
        \footnotesize
        \item[1] Schemes include early (E), intermediate (I) and late (L) collaboration.
        \item[2] \textbf{Data Fusion}: data fusion includes raw data fusion (\textbf{Raw}) and output fusion (\textbf{Out}).
        \item[3] \textbf{Comm Mecha}: communication mechanism includes agent selection (\textbf{Agent}) and feature selection (\textbf{Feat}).
        \item[4] \textbf{Feat Fusion}: feature fusion can be divided into traditional (\textbf{Trad}), graph-based (\textbf{Graph}) and attention-based (\textbf{Atten}) feature fusion.
        \item[5] \textbf{Loss Func}: loss function can be used for uncertainty estimation (\textbf{Uncert}) and redundancy minimization (\textbf{Redund}).  
    \end{tablenotes}
    \end{threeparttable}
    \label{table:methods1}
    \end{table*}

\subsubsection{Customized Loss Function}
Though V2V communication provides a relatively rich perceptual field of view for the ego vehicle, the redundancy and uncertainty of shared information bring new challenges. To overcome these issues, it is important to provide guidance and design specific loss functions for model training.

In the collaboration scene, similar information provided by neighbor agents is redundant to the ego vehicle. To utilize collaborative information effectively, Luo et al. \cite{CRCNet} propose a complementarity-enhanced and redundancy-minimized collaboration network (CRCNet). 
Specifically, CRCNet has two modules to guide the network. 
In the complementarity enhancement module, CRCNet leverages contrastive learning to enhance the information gain. 
In the redundancy minimization module, CRCNet utilizes mutual information to encourage dependence in fused feature pairs. 
With the guidance of the above modules, CRCNet has the ability to select complementary information from neighbor agents when fusing features.

Besides redundancy, the collaborative information also contains perceptual uncertainty, which reflects perception inaccuracy or sensor noises. Su et al. \cite{su2022uncertainty} firstly explore the uncertainty in collaboration perception. Specifically, they design a tailored moving block bootstrap method to estimate the model and data uncertainty, together with a well-designed loss function to capture the data uncertainty directly. Experiments reveal that uncertainty estimation could reduce uncertainty and improve accuracy in different collaboration schemes.

\subsubsection{Output Fusion}

Late collaboration usually adopts fusion operation at the postprocessing stage, which merges multi-agent perception outputs. For example, late collaboration for 3D object detection usually leverages postprocessing methods such as Non-Maximum Suppression (NMS) \cite{NMS} to remove redundant and low-confidence prediction. 
However, late fusion always faces challenges such as spatial and temporal misalignment. Some works \cite{song2023cooperative, DAIR-V2X} propose more robust postprocessing strategies to refine late fusion methods, which will be discussed in the next section.


\subsection{Methods for Real-world Issues} \label{robustness}


\begin{figure*}[htbp]
    \centering
    \includegraphics[width=0.8\linewidth,scale=1.00]{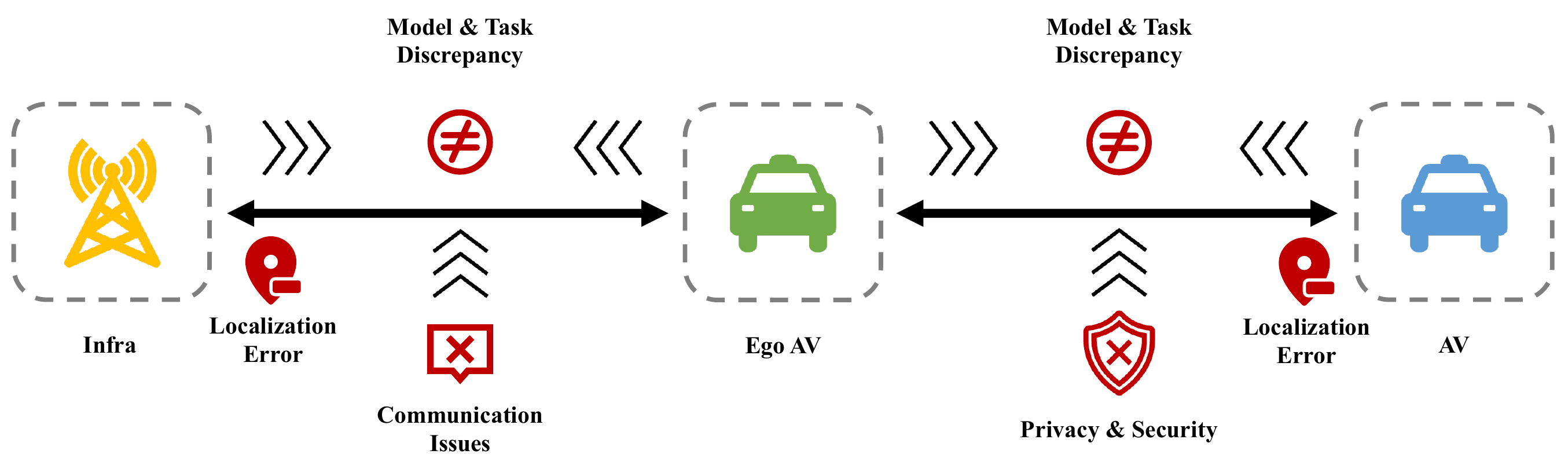}
    \caption{
        Illustration of the collaboration issues in realistic scenarios. Collaborative autonomous vehicles can encounter problems in real applications, such as GPS-induced localization errors, communication latency, model or task discrepancies, and privacy and security issues. Multiple works aim to address these issues and ensure collaboration robustness and safety. }
    \label{fig3:robustness}
\end{figure*}

Most previous collaborative perception research \cite{fcooper,v2vnet, DiscoNet} focuses on collaboration efficiency and perception performance, but all of these methods assume perfect conditions. In real-world autonomous driving scenarios, the communication system may suffer from issues such as 1) localization error, 2) communication latency and interruption, 3) model or task discrepancies and 4) privacy and security issues, as shown in Fig. \ref{fig3:robustness}. With these issues, the collaborative perception system may be damaged, and its performance may be worse than that of individual perception, which seriously threatens the safety of autonomous driving. Therefore, it is of great practical significance to ensure the robustness and safety of collaboration perception. In this subsection, we focus on advanced methods for addressing these issues. The details are as follows.

\subsubsection{Localization Error}

As discussed in the collaboration scheme, collaborative perception methods rely on spatial transformation, which is used to transform raw data, features or outputs. However, GPS localization noises and the asynchronous sensor measurements of agents can introduce localization errors, resulting in data misalignment during aggregation and significant performance degradation in collaborative perception. For example, experiments have shown that V2VNet \cite{v2vnet} is quite vulnerable to pose noise. Recent works propose various pose consistency modules to address this issue.

To tackle localization error issues in V2VNet, Vadivelu et al. \cite{RobustV2VNet} introduce end-to-end learnable neural reasoning layers to correct pose errors. Specifically, Vadivelu et al. \cite{RobustV2VNet} propose a pose regression module and a consistency module before feature aggregation. The pose regression module learns a correction parameter, which will be applied to the noisy relative transformation to produce a predicted true relative transformation. The consistency module refines the predicted relative pose by finding a global consistent absolute pose among all agents with Markov random field (MRF). 

FPV-RCNN \cite{FPV-RCNN} also proposes an effective localization error correction module to avoid performance reduction under localization error. It selects keypoints of poles, fences and walls based on the classification score and utilizes the maximum consensus algorithm with a rough searching resolution \cite{chin2017maximum} to find the corresponding vehicle centers and pole points. Finally, it utilizes these correspondences to estimate pose error. Experiments demonstrate that FPV-RCNN performs better than traditional BEV-based collaboration methods with localization errors.

Late collaboration methods usually adopt straightforward fusion strategies. Accordingly, they are more sensitive to localization errors. To realize robust object-level information combinations, Song et al. \cite{song2023cooperative} design a distributed object-level cooperative perception system called OptiMatch, which utilizes an optimal transport theory-based algorithm to explore fine matching between objects. After the refinement by the above matching algorithms, the late collaboration framework gets a relatively accurate performance even under high location and heading noises.

Similar to OptiMatch \cite{song2023cooperative} that explores pose consistency with object matching algorithms, Lu et al. \cite{coalign} propose a hybrid collaboration framework CoAlign to estimate the correct poses. Specifically, CoAlign constructs an agent-object pose graph where the object nodes are spatially clustering based on the box’s uncertainty estimation. The object pose is sampled from the bounding box cluster, and a pose-consistency optimization function is introduced. Without any pose supervision, CoAlign improves the performance of the collaboration perception network at various noise levels.

\subsubsection{Communication Issues}
Collaborative perception in autonomous driving relies on wireless communication among agents. However, communication may suffer from issues such as latency, interruption and information loss, which will affect the effectiveness of collaboration. In recent years, several efforts \cite{V2X-VIT, SyncNet, DAIR-V2X, LCRN} have been made to explore solutions to these problems. 

To tackle the latency issue in late collaboration, Yu et al. \cite{DAIR-V2X} propose a Time Compensention Late Fusion (TCLF) framework based on tracking and state estimation module. TCLF predicts the current infrastructure prediction with the previous adjacent frame. By matching predictions of the adjacent frame, TCLF can estimate the object velocity and further approximate the object positions at the current frame by linear interpolation. Finally, the estimated infrastructure predictions will be fused with ego prediction.

Compared with TCLF \cite{DAIR-V2X}, V2X-ViT \cite{V2X-VIT} mitigates latency in intermediate collaboration. In particular, V2X-ViT leverages an adaptive delay-aware positional encoding module (DPE) to align features temporally. Moreover, the HMSA and MSwin modules capture inter and intra-agents interactions, which can implicitly correct feature misalignment caused by localization error and time delay. Experiments show that DPE can improve performance under various time delays.

In the same year, Lei et al. \cite{SyncNet} propose the first latency-aware collaborative perception system SyncNet, which realizes a feature-level synchronization. Since features and attention are influenced by each other, the core module SyncNet leverages historical collaboration information to simultaneously estimate the current feature and the corresponding collaboration attention.
Specifically, in the feature-attention symbiotic estimation (FASE) module, dual branches share the same input, which contains real-time and historical features, and learn interactions from previous features/attention, and then estimated features/attention in turn. Furthermore, time modulation adaptively fuses the raw feature and estimated feature based on latency time. 

In addition to latency issue, Ren et al. \cite{V2X-INCOP} firstly consider the communication interruption in collaborative perception. To alleviate the effect, Ren et al. \cite{V2X-INCOP} leverage the historical information to recover missing features and propose an interruption-aware robust collaborative perception (V2X-INCOP) framework. Moreover, they introduce a spatial attention mask to suppress background noise and adopt a curriculum learning strategy to stabilize training.

Packet loss is another critical problem in communication, which may be caused by obstacles and fast-moving vehicles. To address this issue, Li et al. \cite{LCRN} propose an LC-aware Repair Network (LCRN) to ensure the robustness of collaborative perception under lossy communication. Inspired by image-denoising architecture, LCRN adopts an encoder-decoder architecture with a repair loss to recover features from other agents. Moreover, Li et al. \cite{LCRN} present an attention-based fusion method to fuse features between inter and intra-agents to eliminate uncertainty in restoration features, which enhances the model's robustness.

\subsubsection{Model \& Task Discrepancies}

Existing multi-agent collaborative perception methods usually learn in a model-specific and task-specific manner, assuming each agent utilizes the same model to predict outputs for a specific perception task. However, homogeneous models in multi-agents are impractical in the real world. And task-specific training leads to task-specific information, hindering collaborative perception's large-scale deployment. Some works \cite{PSA,csc} provide solutions from a new perspective to mitigate the effect of discrepancies.

When distinct agents are equipped with perception models in different architectures and parameters, current collaboration methods may generate unreliable fusion results due to model heterogeneity. To alleviate this issue, Chen et al. \cite{PSA} propose a model-agnostic collaborative perception framework. Firstly, considering there is a confidence distribution between agents, an offline calibrator is used to align the confidence score of agents to its empirical accuracy. Additionally, to collaborate with the spatial correlation, Chen et al. \cite{PSA} present the Promotion-Suppression Aggregation (PSA) module to find promotion proposals. With offline confidence calibration and online proposal aggregation, the late collaboration method achieves robust performance under parameters and model discrepancy.

Besides the confidence gap in detection bounding boxes, model discrepancy also causes a domain gap between intermediate features. Xu et al. propose the first Multi-agent Perception Domain Adaption framework (MPDA) to address dominant distinctions in intermediate collaboration. Specifically, MPDA resizes the received features to the target size with a Learnable Resizer. It also utilizes a sparse cross-domain transformer to generate domain-invariant features by adversarially fooling a domain classifier. Experiments results demonstrate MDPA effectively bridges the feature domain gap under model discrepancy.

Task-specific training tends to learn task-specific feature representation. The deployment of autonomous driving collaborative perception involves multiple tasks. Therefore the model needs to capture general and robust features. To this end, Li et al. \cite{csc} propose a novel self-supervised learning task termed multi-robot scene completion, which enables each agent to reconstruct a single scene separately to learn latent features. Specifically, they design a spatiotemporal autoencoder (STAR) module to balance the scene reconstruction performance and communication volume in this task. The model learns more robust representations for multi-tasks with the novel autoencoder.

\subsubsection{{Privacy \& Security}}
In addition to the above application issues in collaborative perception, data privacy and model security are crucial. On the one hand, the collaboration training will cause leaks of privacy. On the other hand, the model may be attacked while transmitting data to other agents. This section introduces research on these issues, such as privacy-preserving solutions \cite{xiong2021toward,bi2022edge,9913215} and defence method against attack \cite{AOMAC}.

In collaborative perception, sensor-captured data sharing may lead to significant privacy concerns. Therefore, it's essential to investigate privacy-preserving collaborative perception. Recent studies have proposed various methods to address these privacy concerns in connected and automated vehicles (CAVs), including privacy-preserving convolutional neural network (P-CNN) \cite{xiong2021toward}, edge-cooperative privacy-preserving point cloud object detection framework (SecPCV) \cite{bi2022edge}, and privacy-preserving object detection framework (P2OD) \cite{9913215}. The effectiveness of these methods is achieved through the use of security-preserving technologies on shared raw data, such as additive secret sharing (ASS) and chaotic encryption. It should be noted that the privacy-preserving function of these methods is evaluated exclusively on the individual perception dataset \cite{kitti}, and further evaluation is needed on the collaborative perception datasets.

Collaborative perception relies on communication among agents, while the shared information may be malicious, and the network in agents is vulnerable to adversarial attacks. By studying adversarial robustness, we can enhance the security of collaborative perception. Until now, only one work \cite{AOMAC} investigates adversarial attacks in collaborative perception.
Tu et al. \cite{AOMAC} evaluate the attack and defence performance on V2VNet \cite{v2vnet}. From the perspective of attack, joint perturbations by attackers result in a stronger attack. From the defence perspective, adversarial training can effectively defend against the attack if the attack model is known. Besides, as the number of collaboration agents increases, the defence ability of the collaboration system is enhanced.

\begin{table*}[]
    \vspace{-1em}
    \renewcommand\arraystretch{1.2} 
    \centering 
    \tiny
    \caption{A summary of state-of-the-art collaborative perception methods for real-world issues.}
    \begin{threeparttable}
    \resizebox{\linewidth}{!}{ 
    \begin{tabular}{cc|cc|cccc|c}
    \toprule
    \textbf{Method}       & \textbf{Venue} & \textbf{Modality} &  \textbf{Scheme} & \textbf{Loc Error} &\textbf{Comm Issue} & \textbf{Discrep} & \textbf{Security} & \textbf{Code}                                   \\
    \midrule
    \midrule
    RobustV2VNet \cite{RobustV2VNet}        & CoRL'20    &  LiDAR    & I      &Loc, Pos     & - & -              &-&  - \\
    \midrule
    \midrule
    AOMAC \cite{AOMAC}  & ICCV'21    &     LiDAR     & I     &-       & -&-    & Attack &     -   \\
    P-CNN \cite{xiong2021toward}  & IoT'21    &     Camera     & E     &-       & -&-    & Privacy &     -   \\
    \midrule
    \midrule
    FPV-RCNN \cite{FPV-RCNN}    & RAL'22   &  LiDAR   & I   & Loc, Pos   & -&-    & - & \href{https://github.com/YuanYunshuang/FPV_RCNN}{Link}                                   \\
    TCLF \cite{DAIR-V2X}      & CVPR'22    &   LiDAR  & L &  - & Laten & -   & - &    \href{https://github.com/AIR-THU/DAIR-V2X}{Link} \\
    V2X-ViT \cite{V2X-VIT}   & ECCV'22   &   LiDAR   & I       & Loc, Pos   & Laten & -   & - & \href{https://github.com/DerrickXuNu/v2x-vit}{Link}                                       \\
    SyncNet \cite{SyncNet}   & ECCV'22   &   LiDAR   & I       & -   & Laten & -   & - & \href{https://github.com/MediaBrain-SJTU/SyncNet}{Link}                                       \\ 
    TaskAgnostic \cite{csc}  & CoRL'22   &    LiDAR  &   I     &-     &   - & Task &- &  \href{https://github.com/coperception/star}{Link} \\
    SecPCV \cite{bi2022edge}  & TITS'22    &     LiDAR     & E     &-       & -&-    & Privacy &     -   \\
    \midrule
    \midrule
    ModelAgnostic \cite{PSA}  & ICRA'23   &    LiDAR  &   L     &-     &   - & Model &- &  \href{https://github.com/DerrickXuNu/model_anostic}{Link}    \\
    MPDA \cite{xu2022bridging}  & ICRA'23   &    LiDAR  &   I     &-     &   - & Model &- &  \href{https://github.com/DerrickXuNu/MPDA}{Link}\\
    CoAlign \cite{coalign}  & ICRA'23   &    LiDAR  &   I, L     & Loc, Pos    &   - & - &- &  \href{https://github.com/yifanlu0227/CoAlign}{Link}\\
    LCRN \cite{LCRN}  & TIV'23   &    LiDAR  &   L     &   -  &  Loss & - &- &     -    \\
    OptiMatch \cite{song2023cooperative}  & IV'23   &    LiDAR  &   L     & Loc, Pos    &   - & - &- &     -    \\
    P2OD \cite{9913215}  & IoT'23    &     Camera     & E     &-       & -&-    & Privacy &     -   \\
    V2X-INCOP \cite{V2X-INCOP}   & arXiv'23   &   LiDAR   & I       & -   &  Inter & -   & - &  -        \\
    \bottomrule
    \end{tabular}
    }
    \begin{tablenotes}
        \footnotesize
        \item[1] Schemes include early (E), intermediate (I) and late (L) collaboration.
        \item[2] \textbf{Loc Error} includes localization (\textbf{Loc}) and pose (\textbf{Pos}) errors.
        \item[3] \textbf{Comm Issue} includes latency (\textbf{Laten}), interruption (\textbf{Inter}) and loss (\textbf{Loss}).
        \item[4] \textbf{Discrep} includes model (\textbf{Model}) and task (\textbf{Task}) discrepancies.
        \item[5]\textbf{Security} includes attack defense (\textbf{Attack}) and privacy protection (\textbf{Privacy}).
    \end{tablenotes}
    \end{threeparttable}
    \label{table:methods2}
    \end{table*}

\section{Datasets and evaluation}

Large open datasets are essential in deep learning. Although there are many mature autonomous driving datasets available, such as KITTI \cite{kitti}, nuScenes \cite{nuscenes} and Waymo \cite{waymo}, they focus on individual perception and cannot meet the demand for collaborative perception. Fortunately, recent advances in large-scale benchmark \cite{OPV2V,V2X-Sim,DAIR-V2X,Xu2023V2V4RealAR} for collaborative perception have accelerated the development of autonomous driving perception tasks like detection, segmentation and tracking. 
To follow this progress, we summarize mainstream collaborative perception datasets and provide quantitative comparisons of perception tasks. Since these collaborative datasets do not provide unified evaluation metrics and experimental results for the motion prediction task, we only select three common perception tasks to evaluate, i.e., 3D object detection, 3D object tracking and BEV semantic segmentation. Besides, methods of solving real-world issues adopt different experimental settings. Thus, only a subset of the experimental results for ideal scenarios are presented in this paper.

\subsection{Large-Scale Public Collaborative Perception Datasets}
We summarize typical collaborative perception datasets in Tab. \ref{table:datasets}, including collection sources, data frames, agent type and number, modality and supported perception tasks. We only consider common tasks, and it should be noted that V2V4Real \cite{Xu2023V2V4RealAR} supports domain adaption and DeepAccident \cite{Wang_2023_DeepAccident} supports accident prediction.

Recently, digital twins and parallel vision have contributed to the research on traffic scenes \cite{zhang2020virtual,zhang2021parallel,li2018paralleleye}.
Due to the difficulty of collecting data in reality, most of the collaborative perception datasets are generated by simulation \cite{V2X-Sim, OPV2V, V2X-VIT}, and only a few are collected from real-world \cite{DAIR-V2X, Xu2023V2V4RealAR}.
In this subsection, we introduce several typical datasets that meet the following conditions: (1) published and open-source, (2) large-scale, and (3) providing benchmarks that can be followed. 
We analyze the characteristics of these datasets to facilitate the researcher's selection. 

\begin{table*}[]
    \renewcommand\arraystretch{1.6} 
    \centering 
    \caption{a summary of large-scale collaborative perception datasets.}
    \begin{threeparttable}
    \resizebox{\linewidth}{!}{
        \begin{tabular}{cccc|ccc|ccc|cccc|c}
            \toprule
            \textbf{Dataset} & \textbf{Venue} &  \textbf{Source} & \textbf{Frame} & \textbf{V2V} & \textbf{V2I} & \textbf{Agents} & \textbf{Camera} & \textbf{LiDAR} & \textbf{Depth} & \textbf{OD} &\textbf{SS} &\textbf{OT} & \textbf{MP} & \textbf{Website}                                    \\
            \midrule
            \midrule
            V2V-Sim \cite{v2vnet}  & ECCV'20  & Simu  &  51K & \checkmark  & -    & 1-7  & - & \checkmark & - & \checkmark& - & -& \checkmark& - \\
            \midrule
            \midrule
            V2X-Sim \cite{V2X-Sim}  & RAL'21  & Simu  &  10k & \checkmark  & \checkmark    & 1-5  & \checkmark & \checkmark & \checkmark & \checkmark & \checkmark & \checkmark&- & \href{https://ai4ce.github.io/V2X-Sim}{Link}                          \\
            \midrule
            \midrule
            OPV2V \cite{OPV2V} & ICRA'22  & Simu & 11K & \checkmark & -   & 1-7 & \checkmark  &\checkmark & - & \checkmark & \checkmark & - & - & \href{https://mobility-lab.seas.ucla.edu/opv2v}{Link}                 \\
            DAIR-V2X-C \cite{DAIR-V2X} & CVPR'22  & Real & 39k  &  - & \checkmark   & 2  & \checkmark  & \checkmark & - & \checkmark & - &-& - & \href{https://thudair.baai.ac.cn/coop-dtest}{Link}                      \\
            V2XSet \cite{V2X-VIT} & ECCV'22  & Simu & 11K & \checkmark & \checkmark  & 2-5  & \checkmark   & \checkmark & - &\checkmark & - & - & - & \href{https://github.com/DerrickXuNu/v2x-vit}{Link}                   \\
            DOLPHINS \cite{DOLPHINS} & ACCV'22  & Simu  & 42k & \checkmark  & \checkmark  & 3  & \checkmark & \checkmark & - & \checkmark & - & - & - & \href{https://dolphins-dataset.net}{Link}       \\
            \midrule
            \midrule
            V2V4Real \cite{Xu2023V2V4RealAR} & CVPR'23  &  Real & 20K & \checkmark  &  - & 2  & \checkmark & \checkmark & - & \checkmark & - & \checkmark & - & \href{https://mobility-lab.seas.ucla.edu/v2v4real/}{Link}       \\
            V2X-Seq \cite{v2x-seq} & CVPR'23  &  Real & 15k & -  &  \checkmark & 2  & \checkmark & \checkmark & - & \checkmark & - & \checkmark & \checkmark & \href{https://thudair.baai.ac.cn/coop-forecast}{Link}       \\
            DeepAccident \cite{Wang_2023_DeepAccident} & arXiv'23  & Simu & 57K & \checkmark  &  \checkmark & 1-5  & \checkmark & \checkmark & - & \checkmark & \checkmark & \checkmark & \checkmark & \href{https://deepaccident.github.io/index.html}{Link}       \\
    \bottomrule
    \end{tabular}
    }
    \begin{tablenotes}
        \footnotesize
        \item[1] Source: simulator (Simu) and real-world (Real).
        \item[2] Frame refers to annotated LiDAR-based cooperative perception frame number.
        \item[3] Supported common perception tasks: 3D object detection (OD), BEV semantic segmentation (SS), 3D object tracking (OT), motion prediction (MP).
    \end{tablenotes}
    \end{threeparttable}
    \label{table:datasets}
\end{table*}

\subsubsection{V2X-Sim}
V2X-Sim \cite{DiscoNet,V2X-Sim} is a comprehensive simulated multi-agent perception dataset. It is generated with traffic simulation SUMO \cite{sumo} and CARLA simulator \cite{carla}, and the data format follows nuScenes \cite{nuscenes}. Equipped with RGB cameras, LiDAR, GPS and IMU, V2X-Sim collects 100 scenes with a total of 10,000 frames, each scene contains 2-5 vehicles. Frames in V2X-Sim are divided into 8,000/1,000/1,000 for training/validation/testing. The benchmark of V2X-Sim supports three crucial perception tasks: detection, tracking and segmentation, it should be noted that all tasks adopt a bird's-eye-view (BEV) representation and generate results in 2D BEV. 

\subsubsection{OPV2V}
OPV2V \cite{OPV2V} is another simulated collaborative perception dataset for V2V communication, which is collected with the co-simulating framework OpenCDA \cite{opencda} and CARLA simulator \cite{carla}. The whole dataset is reproducible with provided configuration files. OPV2V contains 11,464 frames with LiDAR points and RGB cameras. A worthy characteristic is that it provides a realistic imitating test set called Culver City, which can be used to evaluate the model's generalization. Its benchmark supports 3D object detection and BEV semantic segmentation, and so far, it only contains one type of object (vehicle).

\subsubsection{V2XSet} V2XSet \cite{V2X-VIT} is a large-scale open simulation dataset for V2X perception. The dataset format is similar to OPV2V \cite{OPV2V} and there are 11,447 frames in total. Compared with the V2X collaboration dataset V2X-Sim \cite{V2X-Sim} and V2I collaboration dataset DAIR-V2X \cite{DAIR-V2X}, V2XSet contains more scenarios, and the benchmark considers imperfect real-world conditions. The benchmark supports 3D object detection and BEV segmentation, with two test settings (perfect and noisy) for evaluation. 

\subsubsection{DAIR-V2X} As the first large-scale V2I collaborative perception dataset from real scenarios, DAIR-V2X \cite{DAIR-V2X} is significant to collaborative perception for autonomous driving. DAIR-V2X-C set can be used to study V2I collaboration, and the VIC3D benchmark is provided to explore V2I object detection tasks. Different from V2X-Sim \cite{V2X-Sim} and V2XSet \cite{V2X-VIT} that mainly focus on LiDAR points, the VIC3D object detection benchmark provides both image-based and LiDAR points-based collaboration methods.

\subsubsection{V2V4Real} V2V4Real \cite{Xu2023V2V4RealAR} is the first large-scale real-world multi-modal dataset for V2V perception, collected with a Tesla vehicle and a Ford Fusion vehicle in Columbus, Ohio, covering 410 km of road. The dataset comprises 20,000 LiDAR frames with over 240,000 3D bounding box annotations for five distinct vehicle classes. Additionally, V2V4Real offers benchmarks for three cooperative perception tasks, including 3D object detection, object tracking, and domain adaptation.

\subsection{Evaluation of Perception Tasks}

\subsubsection{3D Object Detection}

\paragraph{Problem Defination} 
Object detection is one of the most fundamental and challenging problems in computer vision, the goal of which is to localize and recognize objects. According to the dimension of the scene, object detection can be divided into 2D object detection and 3D object detection. Collaborative object detection mainly focuses on 3D object detection. Given a single frame, 3D collaborative object detection models will predict 3D bounding boxes or BEV bounding boxes of the target objects.

\paragraph{Evaluation Metrics} 
Most object detection benchmarks adopt average precision (AP) \cite{voc,kitti,nuscenes} at a specific Intersection-over-Union (IoU) threshold as an evaluation metric.
Average precision (AP) is defined as the area under the continuous precision-recall (PR) curve. In practice, AP is computed through approximate numeric integration over a finite number of sample points, and the mean average precision (mAP) is the average AP of each class. 
Based on the format of model output, the metrics for collaborative detection usually contain $AP_{3D}$ and $AP_{BEV}$, which represent Average Precision (AP) at specific Intersection-over-Union (IoU) thresholds for 3D bounding boxes and 2D bird's eye view (BEV) maps respectively.

\paragraph{Quantitative Results}
To evaluate typical collaboration approaches, we summarize qualitative results of object detection on five dataset benchmarks, and all results are collected from papers and official websites. 
Experiments in Tab.\ref{table:v2x-sim}, Tab.\ref{table:opv2v} and Tab. \ref{table:v2v4real} are in V2V mode, while Tab.\ref{table:dair} and Tab.\ref{table:v2xset} are in V2I and V2X mode, respectively.

In the V2X-Sim BEV detection benchmark (Tab.\ref{table:v2x-sim}), we usually regard early and late collaboration as the upper-bound and lower-bound respectively. Although Who2com \cite{who2com} and When2com \cite{when2com} adopt attention mechanisms to select interacting agents and time, their simple fusion operation makes their performance worse than late collaboration. On the contrary, adaptive fusion strategy-based methods such as V2VNet \cite{v2vnet}, DiscoNet \cite{DiscoNet} and CRCNet \cite{CRCNet} can achieve more ideal results than late collaboration. 

Intermediate collaboration could surpass early collaboration on OPV2V (Tab.\ref{table:opv2v}) and V2XSet (Tab.\ref{table:v2xset}) 3D object detection benchmarks. For the OPV2V benchmark, V2VNet \cite{v2vnet} and FPV-RCNN \cite{FPV-RCNN} perform well based on their ingenious fusion strategy. 
However, in the V2XSet benchmark, V2VNet \cite{v2vnet} and AttFusion \cite{OPV2V} are weaker than F-Cooper \cite{fcooper}, which adopts maxout to fuse features. The possible reason is that a homogeneous collaboration structure brings more noise to the heterogeneous scenario. Especially, V2X-ViT \cite{V2X-VIT} achieves better performance for its specially designed framework for V2X heterogeneous collaboration. 

On DAIR-V2X \cite{DAIR-V2X} dataset, only early and late collaboration results are provided, and both of them improve the performance of vehicle effectively. However, the performance of collaborative perception degrades in the asynchronous phenomenon, TCLF \cite{DAIR-V2X} alleviates the impact of asynchrony to a certain extent by predicting the positions of vehicles with historical information.

V2V4Real \cite{Xu2023V2V4RealAR} evaluates 3D object detection under synchrony and asynchrony settings. As shown in Tab. \ref{table:v2v4real}, collaboration improves detection performance in two settings, and the collaborative detection model has a certain performance degradation in the async scenario. Nonetheless, V2X-ViT \cite{V2X-VIT} and CoBEVT \cite{CoBEVT} achieve relatively stable performance in the async scenario due to well-designed Transformer-based fusion modules.


\begin{table}[]
    \Huge
    \renewcommand\arraystretch{1.4} 
    \centering 
    \caption{Results of BEV object detection ON V2X-Sim \cite{V2X-Sim}.}
    \resizebox{\linewidth}{!}{
    \begin{tabular}{cccccc}
    \toprule
    \textbf{Method}     & \textbf{Modality} & \textbf{Backbone} &\textbf{Scheme} & \textbf{$AP_{BEV}$@0.5} & \textbf{$AP_{BEV}$@0.7} \\
    \midrule
    Individual          &LiDAR & FaF \cite{FaF} & No                            & 45.8            & 40.6            \\
    Late Collab  &LiDAR & FaF \cite{FaF} & Late                          & 55.4            & 41.8            \\
    Early Collab &LiDAR & FaF \cite{FaF} & Early                         & 64.2            & 60.3            \\
    \midrule
    Who2com \cite{who2com}   &LiDAR & FaF \cite{FaF} & Inter                  & 47.2            & 42.2            \\
    When2com \cite{when2com}   &LiDAR & FaF \cite{FaF} & Inter                 & 47.9            & 42.9           \\
    V2VNet  \cite{v2vnet}     &LiDAR & FaF \cite{FaF} & Inter                  & 57.0             & 49.1            \\
    DiscoNet \cite{DiscoNet}   &LiDAR & FaF \cite{FaF} & Inter                  & 60.2            & 53.7            \\
    CRCNet  \cite{CRCNet}    &LiDAR & FaF \cite{FaF} & Inter                 & 61.1            & 55.3            \\
    \bottomrule
    \end{tabular}
    }
    \label{table:v2x-sim}
\end{table}


\begin{table}[]
    \Huge
    \renewcommand\arraystretch{1.4} 
    \centering 
    \caption{Results of 3D object detection ON OPV2V \cite{OPV2V}.}
    \resizebox{\linewidth}{!}{
    \begin{tabular}{cccccc}
    \toprule
    \multirow{2}{*}{\textbf{Method}}  & \multirow{2}{*}{\textbf{Modality}} & \multirow{2}{*}{\textbf{Backbone}} &\multirow{2}{*}{\textbf{Scheme}}  & \multicolumn{2}{c}{\textbf{$AP_{3D}$@0.7}}  \\
                                 &        &            &          & \textbf{Default} & \textbf{Culver} \\
    \midrule
    Individual          &LiDAR & PointPillars \cite{pointpillars}  & No      & 60.2                          & 47.1                        \\
    Late Collab  &LiDAR & PointPillars \cite{pointpillars} & Late   & 78.1                          & 66.8                        \\
    Early Collab &LiDAR & PointPillars \cite{pointpillars} & Early  & 80.0                          & 69.6                        \\
    \midrule
    F-Cooper \cite{fcooper}    &LiDAR & PointPillars \cite{pointpillars} & Inter  & 79.0                            & 72.8                        \\
    V2VNet \cite{v2vnet}   &LiDAR  & PointPillars \cite{pointpillars} & Inter & 82.2                          & 73.4                        \\
    AttFusion \cite{OPV2V}   &LiDAR & PointPillars \cite{pointpillars} & Inter  & 81.5                          & 73.5                        \\
    FPV-RCNN \cite{FPV-RCNN}  &LiDAR  & PV-RCNN \cite{PV-RCNN} & Inter    & 82.0                            & 76.3                       \\
    CoBEVT \cite{CoBEVT}  &LiDAR  & PointPillars \cite{pointpillars} & Inter    & 83.6                            & 73.0                       \\
    \bottomrule
    \end{tabular}
    }
    \label{table:opv2v}
\end{table}

\begin{table}[]
    \Huge
    \renewcommand\arraystretch{1.4} 
    \centering 
    \caption{Results of 3D object detection ON V2XSet \cite{V2X-VIT}.}
    \resizebox{\linewidth}{!}{
    \begin{tabular}{cccccc}
    \toprule
    \multirow{2}{*}{\textbf{Method}} & \multirow{2}{*}{\textbf{Modality}} & \multirow{2}{*}{\textbf{Backbone}} & \multirow{2}{*}{\textbf{Scheme}} & \multicolumn{2}{c}{\textbf{$AP_{3D}$@0.7}} \\
                                     &                                    &                                    &                                                & \textbf{Perfect}  & \textbf{Noisy} \\
    \midrule
    Individual                       & LiDAR                              & PointPillars \cite{pointpillars}                      & No                                             & 40.2             & 40.2          \\
    Late Collab               & LiDAR                              & PointPillars \cite{pointpillars}                      & Late                                           & 62.0              & 30.7          \\
    Early Collab              & LiDAR                              & PointPillars \cite{pointpillars}                      & Early                                          & 71.0              & 38.4          \\
    \midrule
    F-Cooper \cite{fcooper}                        & LiDAR                              & PointPillars \cite{pointpillars}                      & Inter                                   & 68.0              & 46.9          \\
    AttFusion \cite{OPV2V}                          & LiDAR                              & PointPillars \cite{pointpillars}                      & Inter                                   & 66.4             & 48.7          \\
    V2VNet \cite{v2vnet}                        & LiDAR                              & PointPillars \cite{pointpillars}                       & Inter                                   & 67.7             & 49.3          \\
    DiscoNet \cite{DiscoNet}                        & LiDAR                              & PointPillars \cite{pointpillars}                      & Inter                                  & 69.5             & 54.1          \\
    CoBEVT \cite{CoBEVT}                        & LiDAR                              & PointPillars \cite{pointpillars}                      & Inter                                  & 66.0             & 54.3          \\
    Where2comm \cite{where2comm}                        & LiDAR                              & PointPillars \cite{pointpillars}                      & Inter                                  & 68.0             & 45.7          \\
    V2X-ViT \cite{V2X-VIT}                         & LiDAR                              & PointPillars \cite{pointpillars}                      & Inter                                   & 71.2             & 61.4         \\
    \bottomrule
    \end{tabular}
    }
    \label{table:v2xset}
\end{table}

\begin{table}[]
    \Huge
    \renewcommand\arraystretch{1.7} 
    \centering 
    \caption{Results of 3D object detection ON DAIR-V2X \cite{DAIR-V2X}.}
    \begin{threeparttable}
    \resizebox{\linewidth}{!}{
    \begin{tabular}{cccccc}
    \toprule
    \textbf{Method} & \textbf{Modality} & \textbf{Backbone} & \textbf{Dataset} & \textbf{$AP_{3D}$@0.5} & \textbf{$AP_{BEV}$@0.5} \\
    \midrule
    Veh-Only        & Image             & ImvoxelNet \cite{imvoxelnet}        & VIC-Sync         & 12.03             & 13.62              \\
    Inf-Only       & Image             & ImvoxelNet \cite{imvoxelnet}       & VIC-Sync         & 19.93             & 25.31              \\
    Late Collab     & Image             & ImvoxelNet \cite{imvoxelnet}       & VIC-Sync         & 26.56             & 37.75              \\
    \midrule
    Veh-Only        & LiDAR             & PointPillars \cite{pointpillars}     & VIC-Sync         & 31.33             & 35.06              \\
    Inf-Only       & LiDAR             & PointPillars \cite{pointpillars}     & VIC-Sync         & 17.62             & 24.40               \\
    Late Collab     & LiDAR             & PointPillars \cite{pointpillars}     & VIC-Sync         & 41.90              & 47.96              \\
    Early Collab    & LiDAR             & PointPillars \cite{pointpillars}     & VIC-Sync         & 50.03             & 53.73              \\
    \midrule
    Late Collab     & LiDAR             & PointPillars \cite{pointpillars}     & VIC-Async-1      & 40.21             & 46.61              \\
    Late Collab     & LiDAR             & PointPillars \cite{pointpillars}     & VIC-Async-2      & 35.29             & 40.65              \\
    Early Collab    & LiDAR             & PointPillars \cite{pointpillars}     & VIC-Async-1      & 47.47             & 51.67              \\
    TCLF \cite{DAIR-V2X}    & LiDAR             & PointPillars \cite{pointpillars}      & VIC-Async-1      & 40.79             & 46.80               \\
    TCLF \cite{DAIR-V2X}    & LiDAR             & PointPillars \cite{pointpillars}     & VIC-Async-2      & 36.72             & 41.67              \\
    \bottomrule
    \end{tabular}
    \label{table:dair}
    }
    \begin{tablenotes}
        \footnotesize 
        \item[1] VIC-Sync represent temporal synchronous status.
    \end{tablenotes}
    \end{threeparttable}
    \vspace{-0.5em}
\end{table}

\begin{table}[]
    \Huge
    \renewcommand\arraystretch{1.4} 
    \centering 
    \caption{Results of 3D object detection ON V2V4Real \cite{Xu2023V2V4RealAR}.}
    \resizebox{\linewidth}{!}{
    \begin{tabular}{cccccc}
    \toprule
    \multirow{2}{*}{\textbf{Method}} & \multirow{2}{*}{\textbf{Modality}} & \multirow{2}{*}{\textbf{Backbone}} & \multirow{2}{*}{\textbf{Scheme}} & \multicolumn{2}{c}{\textbf{$AP_{3D}$@0.7}} \\
                                     &                                    &                                    &                                                & \textbf{Sync}  & \textbf{Async} \\
    \midrule
    Individual            & LiDAR   & PointPillars \cite{pointpillars}         & No         & 22.0             & 22.0          \\
    Late Collab               & LiDAR        & PointPillars \cite{pointpillars}      & Late         & 26.7              & 22.4          \\
    Early Collab              & LiDAR        & PointPillars \cite{pointpillars}     & Early         & 32.1       & 25.8          \\
    \midrule
    F-Cooper \cite{fcooper}    & LiDAR    & PointPillars \cite{pointpillars}       & Inter       & 31.8              & 26.7          \\
    V2VNet \cite{v2vnet}    & LiDAR      & PointPillars \cite{pointpillars}       & Inter       & 34.3             & 28.5          \\
    AttFusion \cite{OPV2V}     & LiDAR    & PointPillars \cite{pointpillars}     & Inter       & 33.6             & 27.5          \\
    V2X-ViT \cite{V2X-VIT}           & LiDAR   & PointPillars \cite{pointpillars}     & Inter     & 36.9             & 29.3         \\
    CoBEVT \cite{CoBEVT}         & LiDAR    & PointPillars \cite{pointpillars}       & Inter       & 36.0     & 29.7          \\
    \bottomrule
    \end{tabular}
    }
    \label{table:v2v4real}
    \vspace{-0.5em}
\end{table}

\begin{table}[]
    \Huge
    \renewcommand\arraystretch{1.4} 
    \centering 
    \caption{Results of 3D multi-object tracking ON V2V4Real \cite{Xu2023V2V4RealAR}.}
    \resizebox{\linewidth}{!}{
    \begin{tabular}{cccccc}
    \toprule
    \textbf{Method}     & \textbf{Modality} & \textbf{Backbone} &\textbf{Scheme} & \textbf{AMOTA} & \textbf{AMOTP} \\
    \midrule
    Individual          &LiDAR & AB3Dmot  \cite{weng20203d}  & No      & 16.1     & 41.6                        \\
    Late Collab  &LiDAR & AB3Dmot  \cite{weng20203d} & Late   & 29.3     & 51.1                       \\
    Early Collab &LiDAR & AB3Dmot  \cite{weng20203d} & Early  & 26.2      & 48.2                        \\
    \midrule
    F-Cooper \cite{fcooper}    &LiDAR & AB3Dmot  \cite{weng20203d} & Inter  & 23.3     & 43.1                        \\
    V2VNet \cite{v2vnet}   &LiDAR  & AB3Dmot  \cite{weng20203d} & Inter & 30.5     & 54.3                      \\
    AttFusion \cite{OPV2V}   &LiDAR & AB3Dmot  \cite{weng20203d} & Inter  & 28.6     & 50.5                        \\
    V2X-ViT \cite{V2X-VIT}  &LiDAR  & AB3Dmot  \cite{weng20203d} & Inter    & 30.9  & 54.3   \\
    CoBEVT \cite{CoBEVT}  &LiDAR  & AB3Dmot  \cite{weng20203d} & Inter    & 32.1   & 55.6   \\
    \bottomrule
    \end{tabular}
    }
    \label{table:track1}
\end{table}

\begin{table}[]
    \Huge
    \renewcommand\arraystretch{1.4} 
    \centering 
    \caption{Results of BEV multi-object tracking ON V2X-Sim \cite{V2X-Sim}.}
    \resizebox{\linewidth}{!}{
    \begin{tabular}{cccccc}
    \toprule
    \textbf{Method}     & \textbf{Modality} & \textbf{Backbone} &\textbf{Scheme} & \textbf{MOTA} & \textbf{MOTP} \\
    \midrule
    Individual          &LiDAR & SORT \cite{SORT} & No   & 35.7            & 84.2            \\
    Late Collab  &LiDAR & SORT \cite{SORT} & Late       & 21.5            & 85.8            \\
    Early Collab &LiDAR & SORT \cite{SORT} & Early     & 58.0            &  85.6           \\
    \midrule
    Who2com \cite{who2com}   &LiDAR & SORT \cite{SORT} & Inter      & 30.2            & 84.9            \\
    When2com \cite{when2com}   &LiDAR & SORT \cite{SORT} & Inter    & 30.2            & 84.9           \\
    V2VNet  \cite{v2vnet}     &LiDAR & SORT \cite{SORT} & Inter     & 55.3             & 85.2            \\
    DiscoNet \cite{DiscoNet}   &LiDAR & SORT \cite{SORT} & Inter    & 56.7            & 86.2            \\
    \bottomrule
    \end{tabular}
    }
    \label{table:track2}
\end{table}

\begin{table}[]
    \Huge
    \renewcommand\arraystretch{1.4} 
    \centering 
    \caption{Results of BEV semantic segmentation on OPV2V \cite{OPV2V}.}
    \begin{threeparttable}
    \resizebox{\linewidth}{!}{
    \begin{tabular}{cccccc}
    \toprule
    \multirow{2}{*}{\textbf{Method}}   & \multirow{2}{*}{\textbf{Modality}} & \multirow{2}{*}{\textbf{Backbone}} & \multicolumn{3}{c}{IoU} \\
      &  &  & \textbf{Vehicle} & \textbf{Terrain} & \textbf{Lane} \\
    \midrule
    Individual        & Image             & CVT \cite{CVT}             & 37.7             & 57.8             & 43.7          \\
    Map Collaboration & Image             & CVT \cite{CVT}               & 45.1             & 60.0             & 44.1          \\
    \midrule
    F-Cooper \cite{fcooper}         & Image             & CVT \cite{CVT}              & 52.5             & 60.4             & 46.5          \\
    AttFusion \cite{OPV2V}        & Image             & CVT \cite{CVT}              & 51.9             & 60.5             & 46.2          \\
    V2VNet \cite{OPV2V}           & Image             & CVT \cite{CVT}              & 53.5             & 60.2             & 47.5          \\
    DiscoNet \cite{DiscoNet}         & Image             & CVT \cite{CVT}               & 52.9             & 60.7             & 45.8          \\
    FuseBEVT \cite{CoBEVT}         & Image             & CVT \cite{CVT}       & 59.0             & 62.1             & 49.2          \\
    CoBEVT \cite{CoBEVT}           & Image             & SinBEVT \cite{CoBEVT}          & 60.4             & 63.0             & 53.0           \\
    \bottomrule
    \end{tabular}
    }
    \begin{tablenotes}
        \footnotesize 
        \item[1] FuseBEVT only uses the multi-agent fusion module.
        \item[2] CoBEVT uses both of multi-view fusion (SinBEVT) \\ and  multi-agent fusion (FuseBEVT) modules.
        \end{tablenotes}
        \end{threeparttable}
    \label{table:seg1}
    \vspace{-0.5em}
\end{table}

\begin{table*}[]
    \renewcommand\arraystretch{1.4} 
    \centering 
    \caption{Results of BEV semantic segmentation on V2X-Sim \cite{V2X-Sim}.}
    \resizebox{0.9\linewidth}{!}{
        \begin{tabular}{ccccccccccc}
            \toprule
            \textbf{Method}     & \textbf{Modality} & \textbf{Backbone} & \textbf{Vehicle} & \textbf{Pedest} & \textbf{Build} & \textbf{Veget} & \textbf{Sidewalk} & \textbf{Terrain} & \textbf{Road} & \textbf{mIoU} \\
            \midrule
            Individual   & Lidar   & U-Net \cite{UNET}  & 45.93   & 20.59    & 25.38   & 35.83   & 42.39   & 47.03   & 65.76   & 36.64    \\
            Late Collab  & Lidar   & U-Net \cite{UNET}  & 47.67   & 10.78    & 25.26   & 39.46   & 48.79   & 50.92   & 70.00   & 38.38   \\
            Early Collab & Lidar   & U-Net \cite{UNET}  & 64.09   & 31.54    & 29.07   & 45.04   & 41.34   & 48.20   & 67.05   & 42.29          \\
            \midrule
            Who2com \cite{who2com}    & Lidar  & U-Net \cite{UNET}  & 47.74  & 19.16   & 26.11  & 39.64  & 33.60  & 35.81  & 56.75   & 33.81          \\
            When2com \cite{when2com}  & Lidar  & U-Net \cite{UNET}  & 47.74  & 19.16   & 26.11  & 39.64  & 33.60  & 35.81  & 56.75   & 33.81          \\
            V2VNet \cite{v2vnet}      & Lidar  & U-Net \cite{UNET}  & 58.42  & 21.99   & 28.58  & 41.42  & 48.33  & 48.51  & 70.02   & 41.11          \\
            DiscoNet \cite{DiscoNet}  & Lidar  & U-Net \cite{UNET}  & 56.98  & 22.02   & 27.36  & 42.50  & 46.98  & 50.22  & 68.62   & 40.84         \\
            \bottomrule    
        \end{tabular}
    }
    \label{table:seg2}
    \vspace{-0.5em}
\end{table*}

\subsubsection{3D Object Tracking}
\paragraph{Problem Defination} 
Unlike object detection, multiple object tracking (MOT) aims to maintain objects' identities and track their location (3D bounding boxes or BEV bounding boxes) across data frames, which is indispensable for the decision-making of autonomous vehicles. MOT mainly contains two approaches: one is detection-based tracking or tracking-by-detection, and the other is joint detection and tracking. Here we mainly focus on detection-based tracking. 

\paragraph{Evaluation Metrics}
We follow \cite{nuscenes, weng20203d} to employ some metrics to evaluate MOT, including the classic multi-object tracking accuracy (MOTA), multi-object tracking precision (MOTP), average multi-object tracking accuracy (AMOTA) and average multi-object tracking precision (AMOTP). Specifically, the AMOTA and AMOTP average MOTA and MOTP across all recall thresholds.

\paragraph{Quantitative Results}
We summarize multi-object tracking results in Tab.\ref{table:track1} and Tab.\ref{table:track2} and select evaluation metrics according to the experimental results provided by V2V4Real \cite{Xu2023V2V4RealAR} and V2X-Sim \cite{V2X-Sim} benchmarks.
In the V2V4Real dataset (Tab.\ref{table:track1}), the intermediate collaborative method CoBEVT \cite{CoBEVT} achieves the best performance for its fused axial attention (FAX) module. In the V2X-Sim dataset (Tab.\ref{table:track2}), DiscoNet \cite{DiscoNet} is superior to other intermediate performances with its matrix-valued weight design and knowledge distillation framework.

\subsubsection{BEV Semantic Segmentation}

\paragraph{Problem Defination} 
BEV semantic segmentation aims to predict a rasterized map with surrounding semantics under the BEV view. Generally, models take LiDAR points and multi-cameras as input to conduct semantic segmentation. In the collaborative perception scene, multiple agents provide information in distinct views, facilitating semantic scene understanding.

\paragraph{Evaluation Metrics}
To accomplish BEV segmentation, collaborative perception datasets require labeling the semantic segmentation in given categories. The common performance metric for this task is Intersection over Union (IoU) between map prediction and ground truth map-view labels.

\paragraph{Quantitative Results}
We list BEV semantic segmentation results in Tab.\ref{table:seg1} and Tab.\ref{table:seg2}. We only select experiments in V2V mode, and the results demonstrate the effectiveness of collaboration on BEV semantic segmentation. 
In the OPV2V dataset (Tab.\ref{table:seg1}), novel collaboration methods \cite{fcooper,v2vnet,DiscoNet} achieve good segmentation results, while their performance is not as good as CoBEVT \cite{CoBEVT}, which is specially designed for multi-view multi-agent fusion. V2X-Sim benchmark (Tab.\ref{table:seg2}) provides BEV segmentation on six categories. We can find that regularly shaped objects are easier be identified, such as vehicles. Besides, V2VNet \cite{v2vnet} and DiscoNet \cite{DiscoNet} are superior to early collaboration in terms of pedestrian, sidewalk and terrain, which shows that the rich semantic information extracted by each agent is more beneficial to the task.

\section{Challenges and opportunities}

This section presents challenges for conducting collaborative perception in real-world autonomous driving and discusses the corresponding future research directions. 


\subsection{Transmission Efficiency in Collaborative Perception}

Transmission and computing of information occupy lots of inference time of collaborative systems. To realize a real-time collaborative perception system, it is necessary to consider the transmission and computing efficiency. However, most of the works focus on computing efficiency, and only a few works \cite{where2comm,FPV-RCNN} consider transmission efficiency. In order to reduce the latency and improve transmission efficiency, feature compression and selection for transmitted data are vital to collaborative perception.

Current methods \cite{where2comm,FPV-RCNN} usually utilize confidence scores to select critical information, which may ignore the areas with low confidence. 
Future works are encouraged to calculate blind and weak perception areas of the ego vehicle by considering data structure and uncertainty.
Another common method to reduce compression pressure is feature compression.
However, compression methods adopted in current collaborative methods may lose lots of important information, and a more dynamic feature compression strategy such as compression by importance should be considered.

\subsection{Collaborative Perception in Complex Scenes}
It is vital to achieving robust and accurate perception performance under various complex and critical scenes in autonomous driving.
Although several large-scale datasets have emerged in recent years, they are primarily designed for common scenarios and fail to cover complex and challenging scenes (such as bad weather, highway and distant or small objects). In these scenes, sensors may be affected by light or distance to produce low-quality data. In addition, there may be severe spatiotemporal inconsistencies between agents due to high-speed movement, which can lead to instability and uncertainty in collaborative perception systems.

In order to construct a more robust system, there is an urgent need to collect collaborative perception data in complex environments (e.g., DeepAccident \cite{Wang_2023_DeepAccident}) and propose well-designed methods for various complex scenarios. Multi-sensor fusion helps compensate for weather and distance's effects on data quality, and virtual point cloud generation \cite{MVP, pointaugmenting} will contribute to predicting long-range objects. Additionally, spatiotemporal data fusion is required to predict the trajectory of objects moving at high speeds. 



\subsection{Federated Learning-based Collaborative Perception}

In collaborative perception, multiple agents exchange data with each other to improve their models. This approach has been studied extensively, but communication overhead and privacy concerns may arise when agents are from distinct manufacturers or platforms. To protect the privacy of the different autonomous devices and prompt the application of collaborative perception, federated learning (FL) provides a feasible solution.

Federated learning is a method for training machine learning models in decentralized environments where data remains local to each device, which has achieved attention on connected automated vehicles (CAVs) in recent years \cite{chellapandi2023survey}. FL-based collaborative learning enables vehicle collaboration by sharing perception models without requiring direct data exchange. In this way, the vehicles can learn from each other and improve their perception capabilities and communication efficiency in a distributed manner while still preserving data privacy. Existing methods of federated learning for CAVs focus on individual perception, and future works are encouraged to investigate the combination of federated learning and privacy-preserving data exchange in collaborative perception.

\subsection{Collaborative Perception with Low Labeling Dependence}

In recent years, collaborative perception has achieved significant progress. However, the training in collaborative perception systems heavily relies on full-labeling large-scale datasets. The annotations are labor-intensive and time-consuming, especially for collaboration systems involving multiple agents, which seriously affects the research of collaborative perception in the real world. Although some methods have been proposed to reduce the model's dependence on labels in 2D \cite{li2022weakly} and 3D vision \cite{liu2022ss3d}, there are few studies on collaborative perception. In order to promote collaborative perception better, it is crucial to reduce the cost of labeling and investigate collaborative perception with low dependence on labeling.

There are two primary directions to reduce the dependence on labeling. One is generalized weakly supervised learning, and the other is domain adaption. Generalized weakly supervised learning includes semi-supervised learning, which requires a combination of labeled and unlabeled data, and labeling incomplete learning, which requires incomplete annotations for each scene. Domain adaption requires fully annotated source domain data and unlabeled target domain data, for example, fully annotated simulated data and unlabeled real-world data. The domain adaption aims to reduce domain discrepancy between two domains and make collaborative perception models generate domain-invariant features. There have been some attempts at domain adaption on collaborative perception \cite{Xu2023V2V4RealAR,howe2021weakly}, but the wealy-supervised collaborative perception still needs to be explored.
\section{Conclusion}
This work presents a survey on collaborative perception in autonomous driving.
We begin with the collaboration scheme. 
Following that, a comprehensive summary of recent collaborative perception methods is presented. 
Specifically, we systematically summarize the collaborative perception methods for ideal scenarios and real-world issues.
There are also large-scale collaborative perception datasets and performance comparisons on these benchmarks.
Finally, we propose new perspectives with respect to the practical implementation issues of collaborative perception applications.

\ifCLASSOPTIONcaptionsoff
  \newpage
\fi

\bibliographystyle{abbrv}

\bibliography{reference}

\vspace{-18 mm}
\begin{IEEEbiography}
    [{\includegraphics[width=1in,height=1.2in,clip,keepaspectratio]{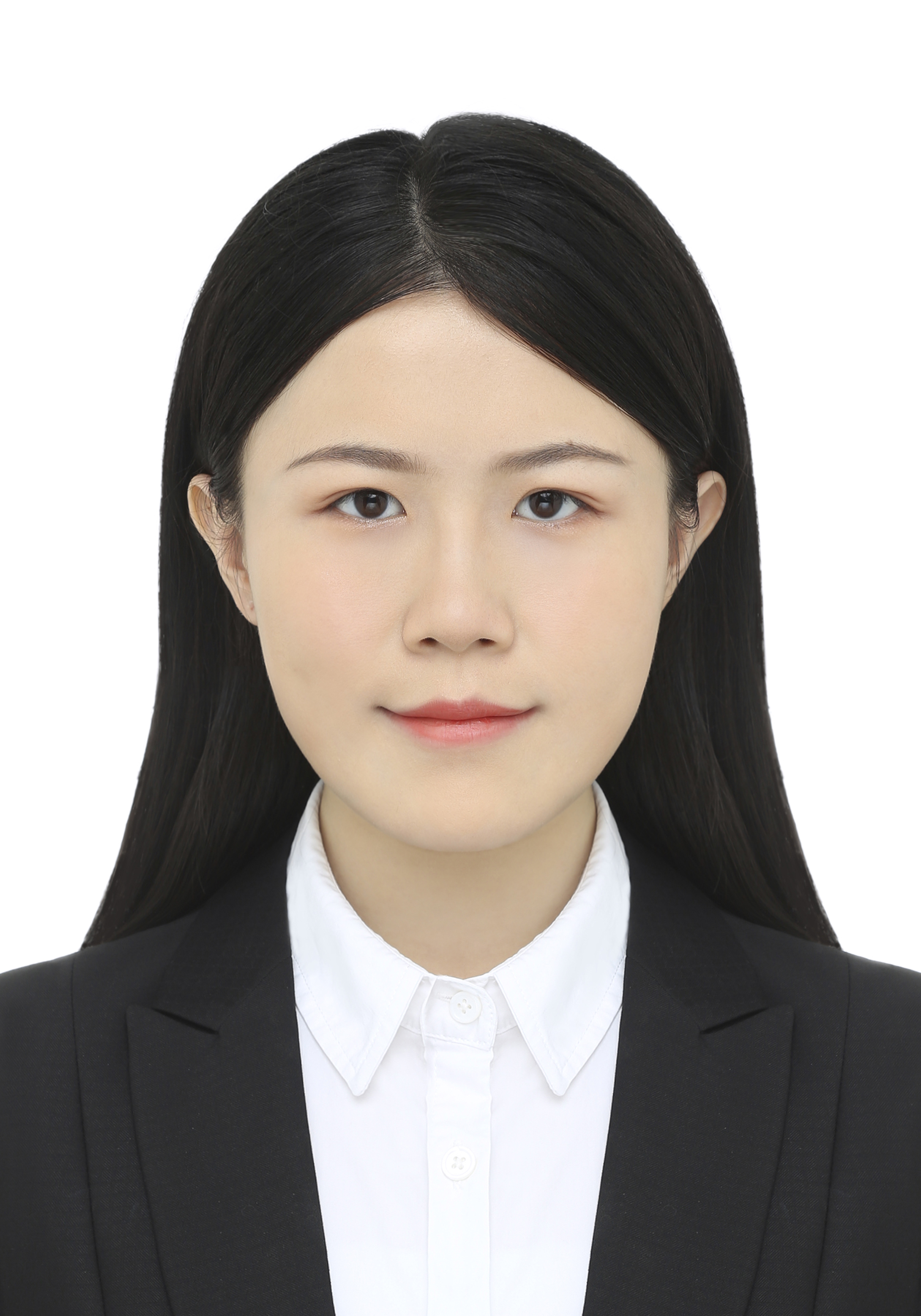}}]{Yushan Han} 
    received the B.S. degree in computer science and technology in 2019 from the School of Computer Science and Information Technology, Beijing Jiaotong University, Beijing, China, where she is currently working toward the Ph.D. degree with the School of Computer Science and Information Technology. Her research interest includes computer vision and autonomous driving.\\
\end{IEEEbiography}

\vspace{-20 mm}
\begin{IEEEbiography}
    [{\includegraphics[width=1in,height=1.2in,clip,keepaspectratio]{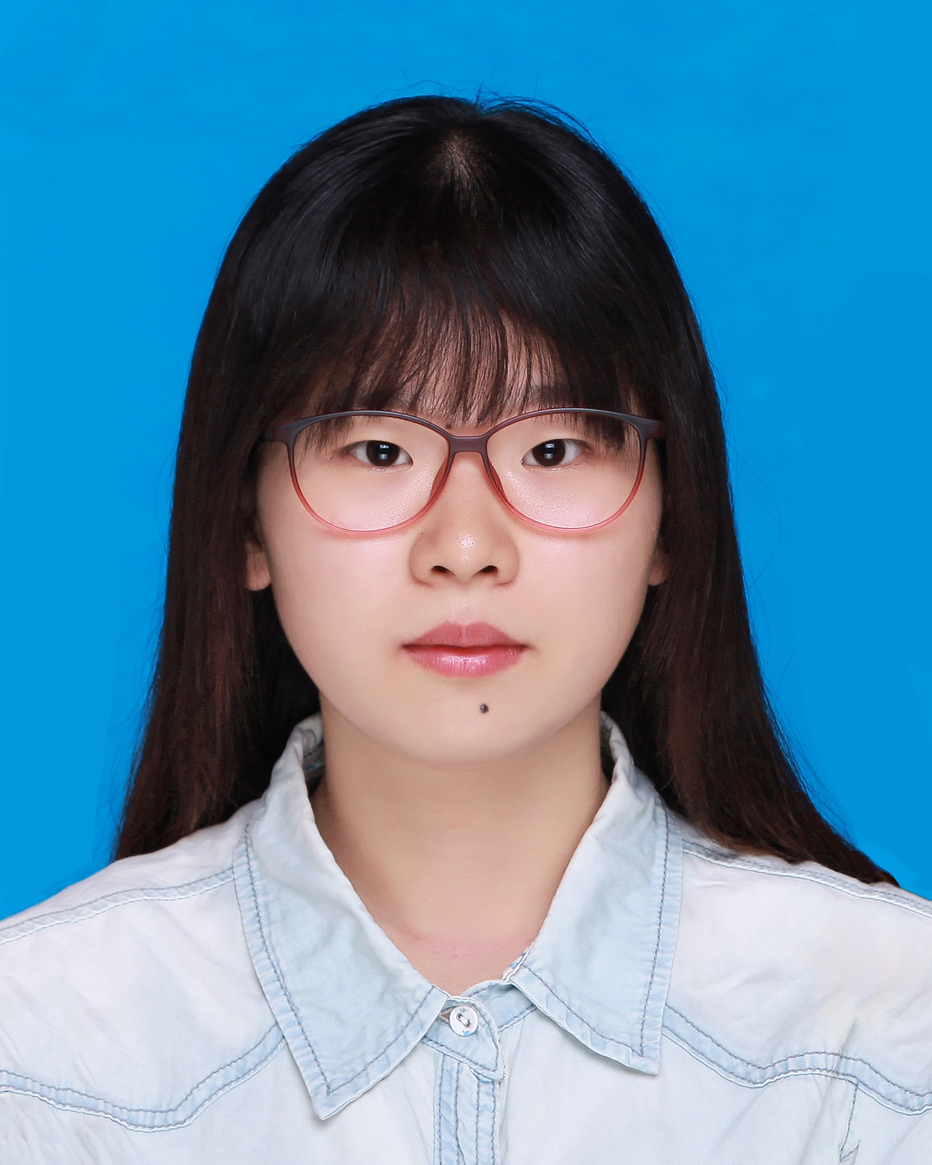}}]{Hui Zhang} (Member, IEEE) received the B.S. degree in automation from the Beijing Jiaotong University, Beijing, China, in 2015 and received the Ph.D. degree in control theory and control engineering from the University of Chinese Academy of Sciences (UCAS), Beijing, China, in 2020. From August 2018 to October 2019, she was supported by UCAS as a joint-supervision Ph.D. student with University of Rhode Island, Kingston, USA. She is currently a lecturer at the School of Computer and Information Technology, Beijing Jiaotong University. Her research interests include computer vision, pattern recognition, and intelligent transportation systems.\\
\end{IEEEbiography}

\vspace{-20 mm}
\begin{IEEEbiography}
    [{\includegraphics[width=1in,height=1.2in,clip,keepaspectratio]{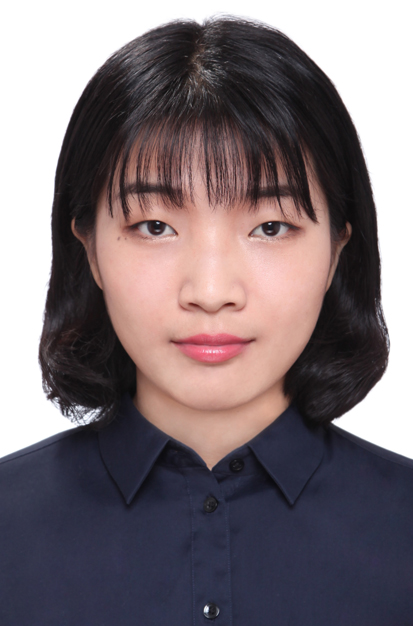}}]{Huifang Li}
    received the M.S. degree in computer science and technology from Beijing Jiaotong University, Beijing, China, in 2017. She received her Ph.D. from the School of Computer and Information Technology at Beijing Jiaotong University, Beijing, China, in 2023. Her research interests include computer vision and machine learning.\\
\end{IEEEbiography}

\vspace{-20 mm}
\begin{IEEEbiography}
    [{\includegraphics[width=1in,height=1.2in,clip,keepaspectratio]{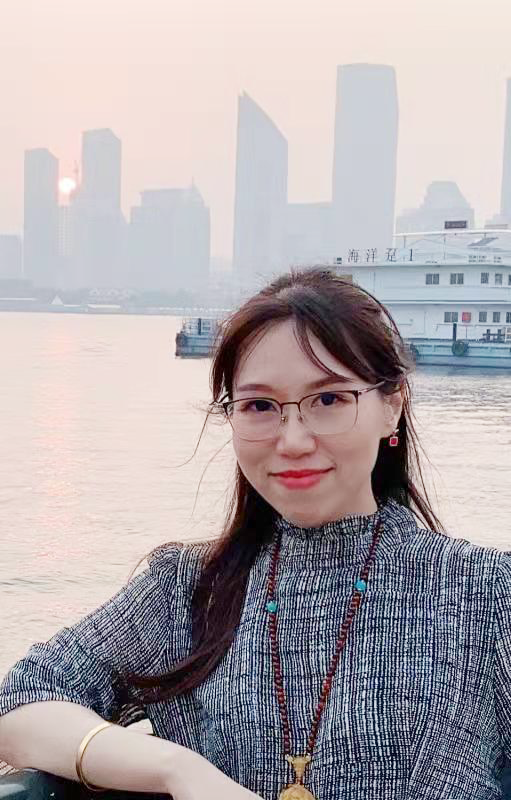}}]{Yi Jin}
    (Member, IEEE) received the Ph.D. degree in signal and information processing from the Institute of Information Science, Beijing Jiaotong University, Beijing, China, in 2010. She was a Visiting Scholar with the School of Electrical and Electronic Engineering, Nanyang Technological University, Singapore, from 2013 to 2014. She is currently a Full Professor with the School of Computer and Information Technology, Beijing Jiaotong University. Her research interests include computer vision, pattern recognition, image processing, and machine learning.\\
\end{IEEEbiography}

\vspace{-20 mm}
\begin{IEEEbiography}
    [{\includegraphics[width=1in,height=1.2in,clip,keepaspectratio]{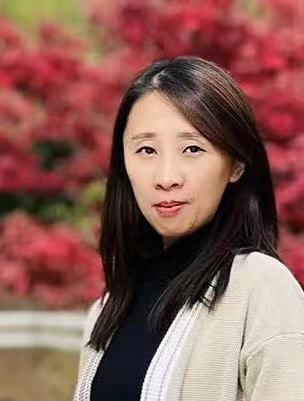}}]{Congyan Lang}
    received the Ph.D. degree from the School of Computer and Information Technology, Beijing Jiaotong University, Beijing, China, in 2006. She was a Visiting Professor with the Department of Electrical and Computer Engineering, National University of Singapore, Singapore, from 2010 to 2011. From 2014 to 2015, she visited the Department of Computer Science, University of Rochester, Rochester, NY , USA, as a Visiting Researcher. She is currently a Professor with the School of Computer and Information Technology, Beijing Jiaotong University. Her current research interests include multimedia information retrieval and analysis, machine learning, and computer vision.\\
\end{IEEEbiography}

\vspace{-20 mm}

\begin{IEEEbiography}
    [{\includegraphics[width=1in,height=1.2in,clip,keepaspectratio]{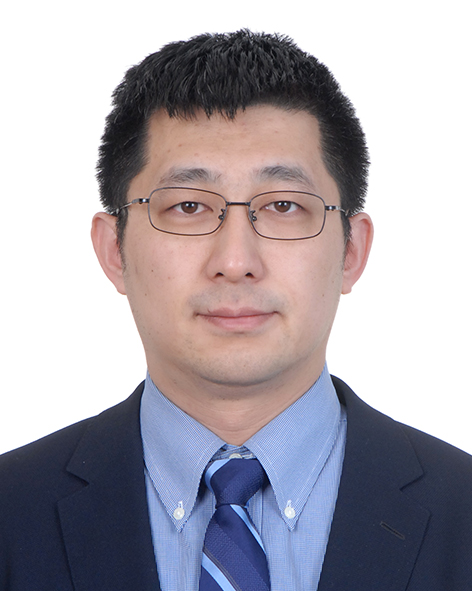}}]{Yidong Li} 
    (Senior Member, IEEE) received the B.Eng. degree in electrical and electronic engineering from Beijing Jiaotong University in 2003, and the M.S. and Ph.D. degrees in computer science from The University of Adelaide, in 2006 and 2010, respectively. He is currently the Vice-Dean and a Professor with the School of Computer and Information Technology, Beijing Jiaotong University. He has published more than 150 research papers in various journals and refereed conferences. His research interests include big data analysis, privacy preserving and information security, data mining, social computing, and intelligent transportation.\\
\end{IEEEbiography}



\end{document}